\documentclass{article}
\usepackage{iclr2027_conference,times}

\usepackage{amsmath,amsfonts,bm}

\def\eqref#1{equation~\ref{#1}}

\def\1{\bm{1}}

\DeclareMathAlphabet{\mathsfit}{\encodingdefault}{\sfdefault}{m}{sl}
\SetMathAlphabet{\mathsfit}{bold}{\encodingdefault}{\sfdefault}{bx}{n}

\usepackage{amsmath,amssymb}
\usepackage{booktabs}
\usepackage{graphicx}
\usepackage{multirow}
\usepackage{subcaption}
\usepackage{tabularx}
\usepackage{xcolor}
\usepackage{tikz}
\usetikzlibrary{arrows.meta,calc,positioning,shapes.geometric}
\usepackage[most]{tcolorbox}
\usepackage{microtype}
\usepackage{wrapfig}
\usepackage{needspace}
\usepackage{url}
\usepackage{hyperref}
\usepackage{marvosym}

\definecolor{undcolor}{HTML}{78669A}
\definecolor{gencolor}{HTML}{9C7BB5}
\definecolor{jointcolor}{HTML}{D95F59}
\definecolor{muted}{HTML}{666666}
\definecolor{findingblue}{HTML}{7F9365}
\definecolor{findingbluebg}{HTML}{EDF0E7}
\newcommand{\und}{\textsc{Und}}
\newcommand{\gen}{\textsc{Gen}}
\newcommand{\joint}{\textsc{Und+Gen}}
\newcommand{\umm}{\textsc{UMM}}
\newcommand{\mot}{\textsc{MoT}}

\newcommand{\findinglabel}[1]{\textcolor{findingblue}{\textbf{#1}}}

\newcommand{\paperfigref}[1]{Figure~\ref{#1}}
\newcommand{\papertabref}[1]{Table~\ref{#1}}
\newtcolorbox{findingbox}{
  enhanced,
  breakable,
  colback=findingbluebg,
  colframe=findingblue,
  boxrule=0.8pt,
  arc=2.5mm,
  left=3mm,
  right=3mm,
  top=2.5mm,
  bottom=2.5mm,
  boxsep=0pt,
  before skip=14pt,
  after skip=8pt,
}
\newcommand{\correspondingauthor}{%
  \textsuperscript{\href{mailto:ziwei.liu@ntu.edu.sg}{\Letter}}}

\title{Uncovering Understanding--Generation Synergy in Native Unified
Multimodal Models:\\
From Representation, Task to System}

\author{
Penghao Wu$^{1}$ \quad Haiwen Diao$^{1}$ \quad Weichen Fan$^{1}$ \quad
Lewei Lu$^{2}$ \quad Dahua Lin$^{2}$ \quad Ziwei Liu$^{1}$\correspondingauthor\\[3pt]
\normalfont $^{1}$S-Lab, Nanyang Technological University \qquad $^{2}$SenseTime Research\\[4pt]
\small\texttt{\{penghao001,weichen002\}@e.ntu.edu.sg}\qquad
\small\texttt{\{haiwen.diao,ziwei.liu\}@ntu.edu.sg}\\
\small\texttt{\{luotto,dhlin\}@sensetime.com}
}

\iclrfinalcopy

\begin{document}
\maketitle
\fancyhead{}
\renewcommand{\headrulewidth}{0pt}

\begin{abstract}
While unified multimodal models (UMMs) jointly perform visual understanding and generation within a single model, functional unification does not guarantee learning synergy: the two objectives may reinforce each other, compete for capacity, or merely coexist. We investigate their relationship at the representation, task,
and system levels in a controlled, structurally native 
setting without pretrained vision priors. At the
representation level, we find that each objective provides useful signal to the other:
generation enriches the visual features learned for understanding, while
understanding strengthens vision--language alignment for generation. However,
when both objectives are forced through the same computation path, one tends
to dominate. A task-decoupled architecture that specializes conflicting visual
computation while preserving semantic interaction avoids this asymmetric
degradation. At the task level, through three case studies, we find positive bidirectional transfer when understanding
and generation tasks rely on shared knowledge. At the system level, we show that an end-to-end UMM outperforms a matched planner--executor pipeline on complex tasks that explicitly require both image understanding and generation. Together, these results show that the value of UMMs extends beyond a unified interface: appropriate specialization, shared task knowledge, and end-to-end optimization can turn coexistence into synergy.
\end{abstract}

\section{Introduction}
\label{sec:intro}
Unified multimodal models (UMMs) bring visual understanding and generation
into a single model, allowing text and images to be consumed and produced as
interleaved content
\citep{chameleon2024,transfusion2024,showo2024,janus2024,bagel2025,liu2025tuna,sensenova2026sensenovau1}. This
capability enables forms of interaction that are difficult to realize with a
conventional understanding-only or generation-only model. For example, UMMs can perform reasoning by generating
mixed visual and textual content and iteratively interpret and modify images in the same sequence.
Such applications provide a compelling functional motivation for unification.

However, the ability to expose understanding and generation through one interface
does not establish that the two capabilities intrinsically benefit one another within the
model. Their objectives may reinforce shared visual representations, compete for
representational and computational capacity, or simply coexist without
meaningful transfer. Existing UMMs primarily demonstrate that both
capabilities can be supported at scale, while offering limited and sometimes
conflicting evidence about how their joint learning changes either capability \citep{kang2026transferability,han2026towards,liu2025tuna,blip3o,metaquery}.
Consequently, a basic question underlying unified multimodal intelligence
remains unresolved: \emph{when and why do visual understanding and generation help
each other, and when does unification instead create interference?}

\begin{figure}[t]
  \centering
  \includegraphics[width=0.9\linewidth]{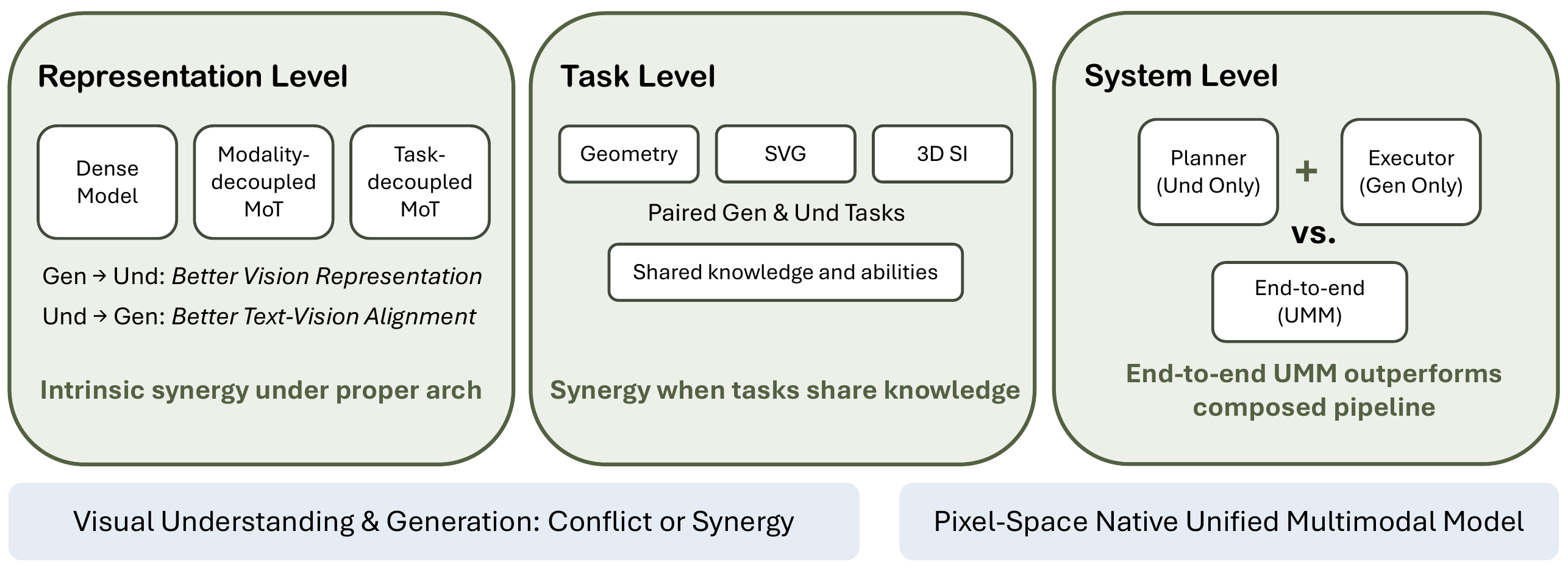}
  \caption{{Overview of our three-level study of
  understanding--generation synergy in native UMMs.} At the representation level, we study
  the intrinsic relationship between understanding and generation under different architectures. At the task
  level, we test whether shared task knowledge enables bidirectional transfer. At
  the system level, we compare an end-to-end UMM with a composed pipeline of comparable understanding and generation capability.}
  \label{fig:teaser}
  \vspace{-0.8\baselineskip}
\end{figure}

To study this question under controlled conditions, we adopt a \textit{native}
multimodal setting in which images are both consumed and produced directly in
pixel space. Raw image patches enter the model without a pretrained vision
encoder, while images are generated without an image tokenizer or VAE \citep{VAE}. This
pixel-in, pixel-out design minimizes external visual priors and allows us to
more cleanly isolate the interaction between understanding and generation. We
use a mature hybrid formulation that combines autoregressive text modeling
with flow matching for visual generation. Starting from the same pretrained
language model, we compare understanding-only, generation-only, and jointly
trained variants under controlled routing and parameter-sharing strategies.

Within this controlled setting, we organize our investigation at three
progressively broader levels. At the \textbf{representation level}, we examine
how understanding and generation jointly shape visual representations, and how
parameter sharing and architectural specialization determine transfer or
interference between their learning objectives. At the \textbf{task level}, we
investigate whether related understanding and generation tasks benefit from
joint training when they rely on shared domain knowledge or capabilities. At
the \textbf{system level}, we study complex tasks that inherently require both
capabilities, directly comparing an end-to-end UMM with an agentic pipeline
composed of separate understanding and generation models.

At the representation level, we first study two different routing settings. In the
\emph{dense shared model}, the same pretrained LLM processes text, clean visual
tokens for understanding, and noised visual tokens for generation. In the
\emph{modality-decoupled} \mot{} \citep{mot2025}, text remains in the pretrained LLM branch,
while all visual tokens are routed through a separate branch initialized from
scratch. The two settings exhibit opposite asymmetries. In the dense model,
joint training improves visual understanding but degrades generation; in the
modality-decoupled \mot{}, it improves generation but degrades understanding.
Despite this trade-off, through representation-level probing, we find that each objective provides useful signal to the other:
generation enriches the visual representations used for understanding, while
understanding better aligns visual and language representations for
generation. The problem
is that forcing both objectives to use the same computation path makes one
dominate and reduces the other largely to a regularizer. Based on this
diagnosis, we develop a \emph{task-decoupled} \mot{}, which keeps understanding
visual tokens in the language branch and assigns generation visual tokens to a
specialized branch, thereby achieving soft-alignment between understanding and generation visual tokens. This
design allows the two capabilities to coexist without the same sacrifice.

At the task level, we study three domains in which understanding and
generation could potentially share common knowledge: geometry reasoning,
vector graphics, and 3D spatial intelligence. For geometry, we pair geometry
problem solving with text-to-image generation and editing of geometric
diagrams. Joint training produces clear gains on geometry reasoning and
further improves performance on general mathematical reasoning benchmarks. In
vector graphics and 3D spatial intelligence, jointly training the paired
understanding and generation tasks improves both directions. Across these
case studies, the same pattern emerges: visual understanding and generation benefit
one another when their tasks require shared knowledge and capabilities.

At the system level, we ask whether an end-to-end UMM is more effective than
connecting separate understanding and generation models for complex tasks that require both image understanding and generation. We study
reasoning-intensive image editing, where the model must understand the source
image and request, reason about the explicit edit instruction, and generate the resulting
image. We compare an end-to-end UMM with a matched agentic pipeline in which an
understanding model first produces an explicit edit instruction and a
generation model then executes it. Under the same base model and data
supervision, the end-to-end UMM shows clear advantages. This result shows that for
complex tasks requiring tight interaction between understanding and
generation, learning the full process within one unified model can be more effective
than composing the two capabilities as separate stages.

Together, these studies show that the benefit of unifying understanding and
generation is conditional rather than automatic. At the representation level,
the architecture determines whether useful transfer becomes mutual improvement
or an asymmetric trade-off. At the task level, joint training is most useful
when the two directions require the same domain knowledge. At the system
level, end-to-end modeling is advantageous when understanding and generation
must interact throughout the task. These findings provide concrete guidance
for building UMMs: specialize computation when the learning objectives
conflict, and unify training and inference when knowledge and decisions are
shared across the two capabilities.

Our contributions are:
\begin{itemize}
  \item We present a controlled and comprehensive study of understanding--generation interaction
  at the representation, task, and system levels under a native multimodal setting.
  \item We reveal that understanding and generation can provide useful learning
  signals to each other, but whether these signals translate into mutual gains
  is architecture-dependent.
  \item We find that task-level synergy emerges when understanding and
  generation rely on shared knowledge or capabilities, through case studies spanning a broad spectrum from geometric reasoning to spatial intelligence.
  \item We establish a system-level advantage of unification for tasks that
  require tightly coupled understanding, reasoning, and generation by showing that an
  end-to-end UMM outperforms a matched planner--executor pipeline on
  reasoning-intensive image editing.
\end{itemize}

\section{Native Unified Modeling}
\label{sec:model}

\subsection{Scope and formulation}

\paragraph{Modeling choice.}
Unified multimodal models span several modeling paradigms. Images and text can
both predicted autoregressively with image being discretized \citep{chameleon2024,januspro,cui2025emu35}
or predicted using a flow-matching head \citep{mingunivision,Harmon}; both modalities can also be trained with a discrete
denoising objective \citep{xin2025lumina,ai2026llada2}; or autoregressive text modeling
can be combined with continuous diffusion or flow matching for images
\citep{transfusion2024,bagel2025}. We adopt the last formulation, which is currently a
widely used and empirically strong choice for jointly supporting
language understanding and high-quality image generation. Specifically, the multimodal context
is modeled causally, while the visual tokens within each
generated image are modeled bidirectionally with flow matching. Evaluating
whether the same interactions hold under fully autoregressive or discrete
denoising formulations is left to future work. We nevertheless expect many of
the conclusions about parameter sharing, task transfer, and system design to
extend beyond the particular generative objective used here.

\paragraph{Native visual interface.}
Our use of \emph{native} in this study refers specifically to how visual information enters
and leaves the model, rather than to multimodal pretraining from scratch. The pixel-in, pixel-out design avoids importing
visual representations learned by external models, which could otherwise mask
whether changes in visual features arise from understanding--generation joint
training or from the pretrained visual prior itself.


\subsection{Two controlled routing settings}
\label{sec:routing}

\begin{figure}[t]
  \centering
  \includegraphics[width=\linewidth]{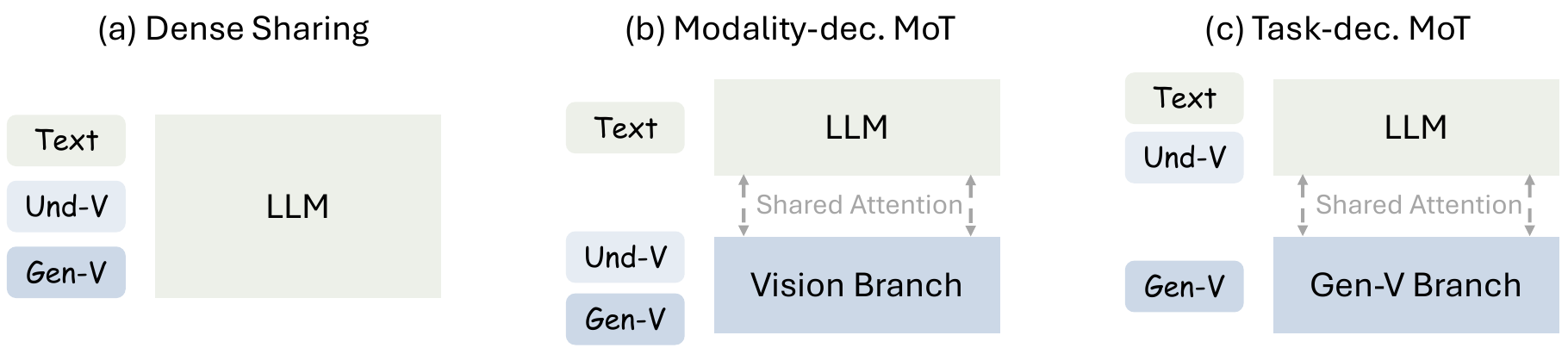}
  \caption{\textbf{Illustration of different architectures in our study.}
  (a) Dense sharing sends all tokens through the same LLM decoder.
  (b) Modality-decoupled \mot{} keeps text in the LLM and sends all visual
  tokens through a scratch-trained branch. (c) Task-decoupled \mot{} keeps \textsc{Und-V}
  language-anchored while specializing \textsc{Gen-V}. For simplicity, we omit the pre-buffer layers here.}
  \label{fig:architectures}
\end{figure}

We start from the pretrained Qwen3-1.7B language model
\citep{qwen3_2025} and augment it with a stack of randomly initialized
pre-buffer layers, following the native multimodal design of
\citet{neo}. Image inputs are first converted into patch tokens by two
convolutional layers with a patch size of 32 and then passed to the LLM decoder
layers. Attention is causal across the overall multimodal sequence, while
tokens within the same image attend bidirectionally to one another. MRoPE-I
\citep{mrope} is used to encode multimodal positions while preserving the 2D
spatial structure of images and the positional priors of the pretrained LLM.
To isolate how access to pretrained
language parameters affects the interaction between understanding and
generation, we compare the two routing settings illustrated in
\paperfigref{fig:architectures}(a--b).

\paragraph{Dense sharing.}
The most direct design sends every token type---text, clean visual tokens for
understanding, and noised visual tokens for generation---through the same LLM layers. This setting maximizes parameter sharing
and gives both types of visual tokens direct access to pretrained language
semantics. We train three matched variants: \und{} uses only understanding
data, \gen{} uses only generation data, and \joint{} uses both.

\paragraph{Modality-decoupled \mot{}.}
To isolate visual representation learning from the pretrained language
backbone, we construct a modality-decoupled variant following the token-routed
principle of \mot{} \citep{mot2025}. The pretrained branch processes text
only, while a parallel branch initialized from scratch processes all visual
tokens, including both understanding tokens (\textsc{Und-V}) and noised
generation tokens (\textsc{Gen-V}). The two branches retain global attention
over the interleaved sequence.

\section{Representation-Level Synergy}
\label{sec:representation}

We first study how joint training reshapes the foundational visual representations. Specifically, we investigate whether parameter sharing intrinsically fosters representational synergy or induces feature interference, and how routing choices affect this dynamic.

\subsection{Training settings and evaluation}

\paragraph{Training protocol.}
We mainly select understanding and generation data from SenseNova-U1 \citep{sensenova2026sensenovau1} for this study. The
understanding data covers diverse types of multimodal instruction tuning data. The generation stream contains text-to-image generation and image
editing data. Detailed training data statistics are reported in
Section \ref{app:representation_details_data}. All variants use single-stage training with 210k optimization steps and a constant learning rate of $1\times10^{-4}$. More training details are provided in
Section~\ref{app:representation_details_training}.

\paragraph{Evaluation.}
We group understanding benchmarks into three categories. \textbf{General}
contains MME \citep{mme}, MMBench \citep{MMBench}, MMStar \citep{mmstar}, SEED Bench \citep{seedbench}, and MMMU \citep{MMMU}. \textbf{OCR} contains DocVQA \citep{docvqa}, ChartQA \citep{chartqa}, InfoVQA \citep{infovqa}, OCRBench \citep{ocrbench}, and
AI2D \citep{ai2d}. \textbf{Vision-centric \& SI (Spatial Intelligence)} contains
PerceptionBench \citep{lin2026perceptionbench}, P2GB \citep{p2gb}, BLINK \citep{fu2024blink}, MME-Realworld \citep{mme-realworld}, DA2K \citep{DA2K}, CV-Bench \citep{tong2024cambrian}, MindCube \citep{mindcube}, 3DSR \citep{ma20253dsrbench}, and ViewSpatial \citep{viewspatial}.
Generation is evaluated on GenEval2 \citep{geneval2}, DPGBench \citep{DPG}, HPSv3 \citep{ma2025hpsv3}, and Aesthetic Score \citep{discus0434_aesthetic_predictor_v2_5}.
Full per-benchmark results and evaluation details are reported in Section \ref{app:representation_details_evaluation}.

\subsection{Results of Dense sharing}

\begin{table}[t]
  \caption{Understanding and generation results across routing architectures.
  Understanding columns report category averages. Atom-level score is reported for GenEval2. Higher is better for all metrics.}
  \label{tab:representation-results}
  \centering
  \setlength{\tabcolsep}{2.5pt}
  \scalebox{0.85}{%
  \begin{tabular}{cllccccccc}
    \toprule
    \multirow{2}{*}{Model} & \multicolumn{2}{c}{Model}
      & \multicolumn{3}{c}{Understanding}
      & \multicolumn{4}{c}{Generation} \\
    \cmidrule(lr){2-3}\cmidrule(lr){4-6}\cmidrule(lr){7-10}
    & Architecture & Training & General & OCR & V-Centric \& SI
      & GenEval2 & DPG  & HPSv3 & Aesthetic \\
    \midrule
    1 & \multirow{2}{*}{Dense} & \und{}/\gen{}
      & 67.82 & 71.96 & 60.72 & 51.16 & 79.11 & 7.05 & 5.12 \\
    2 & & \joint{} & 68.69 & 72.12 & 61.77
      & 51.03 & 78.12 & 5.92 & 4.94 \\
    \midrule
    3 & \multirow{2}{*}{Modality-dec. \mot{}} & \und{}/\gen{}
      & 64.20 & 66.84 & 59.54 & 57.55 & 82.06  & 7.72 & 5.37 \\
    4 & & \joint{} & 61.58 & 62.11 & 57.49
      & 63.47 & 83.09 & 7.97 & 5.48 \\
    \midrule
    5 & Task-dec. \mot{} & \joint{} & 69.02 & 72.09 & 62.38
      & 63.96 & 82.31 & 7.90 & 5.50 \\
    \bottomrule
  \end{tabular}
  }
\end{table}

As shown in \papertabref{tab:representation-results} (Rows 1--2), for visual understanding, joint training achieves better performance compared with the \und{} baseline, with the
largest gain on Vision-centric \& SI. The effect is nevertheless
strongly asymmetric: compared with the \gen{} model, the \joint{} model
performs substantially worse across the generation metrics. 

\begin{figure}[t]
  \centering
  \includegraphics[width=\linewidth]{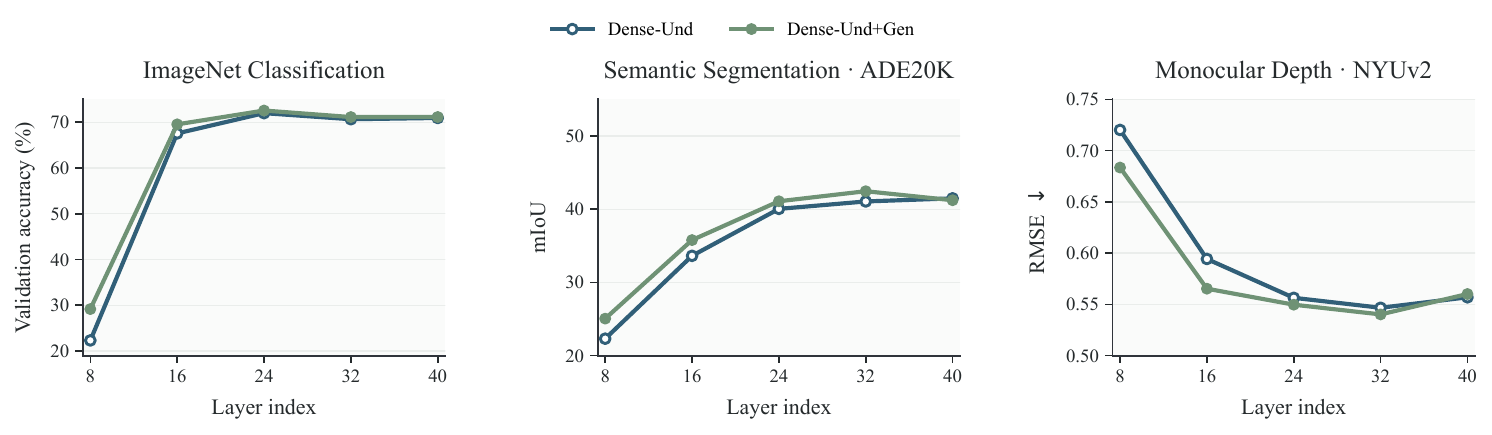}
  \caption{Layer-wise probing experiments of visual representations. Higher is better for
  classification and mIoU; lower is better for depth R-MSE. Joint training strengthens visual representations, with the clearest gains in early and middle layers across semantic and geometric probing tasks.} 
  \label{fig:dense-probes}
\end{figure}

\begin{figure*}[h]
  \centering
  \includegraphics[width=\textwidth]{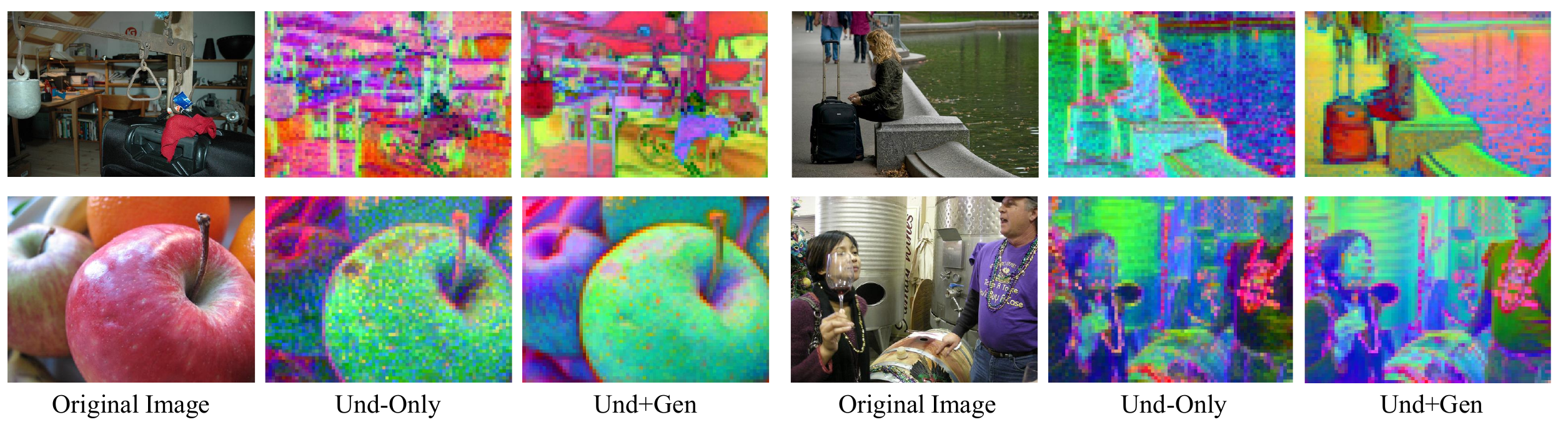}
  \caption{\textbf{PCA visualization of dense visual features.}
  Patch-level features are projected onto a shared three-dimensional PCA basis
  and visualized as RGB maps. Compared with understanding-only training, joint
  training produces more coherent object regions and clearer spatial
  structure, consistent with the improvements observed by the frozen probes.}
  \label{fig:pca-visualization}
\end{figure*}

To understand how joint training changes the visual representations for visual understanding, we conduct a series of layer-wise probes on the frozen
backbones. We measure global semantic information with ImageNet \citep{ImageNet}
classification and dense spatial information with semantic segmentation on ADE20K \citep{ade20k} and
monocular depth estimation on NYU-Depth V2 \citep{nyudepthv2}. We additionally visualize PCA projections of the
visual features for qualitative comparison. From the probing results (\paperfigref{fig:dense-probes}) and feature visualization (\paperfigref{fig:pca-visualization}), we observe that the generation supervision improves the visual representations learned for visual understanding.

Recent work shows that image and video generation training can learn strong,
general-purpose visual representations, supporting zero-shot or data-efficient
transfer to a broad range of visual understanding tasks
\citep{gabeur2026imagegenerators,wang2026videogeneration}.Our probing results
are consistent with this broader observation: generation supervision enriches
the visual features for visual understanding. The dense setting, however,
reveals that better representations for understanding do not automatically yield
balanced understanding and generation within a unified model. We hypothesize that the pretrained
LLM acts as a strong semantic anchor, allowing understanding to dominate the
shared feature space. Generation consequently serves as a useful auxiliary
constraint for representation learning, but its own capability is degraded by
competition within the shared parameters.

\Needspace{0.42\textheight}
\subsection{Results of Modality-decoupled \mot{}}

\begin{wrapfigure}[16]{r}{0.36\linewidth}
  \centering
  \vspace{-0.6\baselineskip}
  \includegraphics[width=\linewidth]{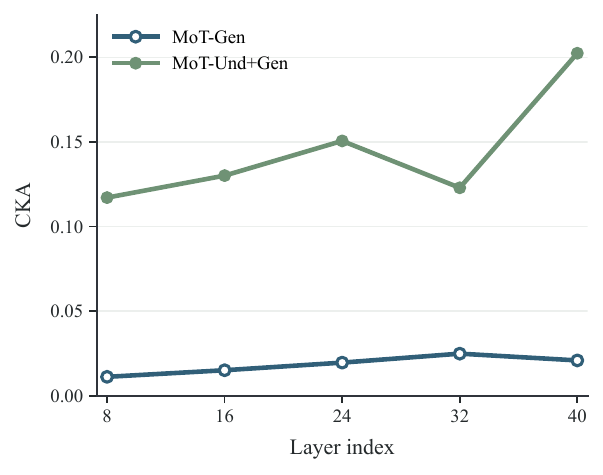}
  \caption{Joint learning improves vision--language alignment for generation.}
  \label{fig:cka-alignment}
  \vspace{-0.6\baselineskip}
\end{wrapfigure}

As shown in \papertabref{tab:representation-results} (Rows 3--4), when all visual tokens move to the scratch-trained visual-only branch, \joint{}
is worse than \und{} on visual understanding benchmarks, while generation performance becomes significantly better than \gen{}. In contrast to the dense architecture, generation dominates the shared visual computation in this setting, while understanding primarily serves as auxiliary supervision. Moreover,
the modality-decoupled \und{} model is below its dense counterpart,
particularly on General and OCR, indicating that, for
tasks that rely on strong vision--text alignment, processing visual and textual
tokens with shared parameters is more effective than separating them into
modality-specific branches.

Since joint training improves generation in this setting, we further examine
whether understanding supervision changes the generative visual
representations. We measure layer-wise linear CKA \citep{cka2019} between
pooled text features from the language branch and generative image features from the visual branch using 5000 text-image pairs from the COCO validation set \citep{coco}. As shown in
\paperfigref{fig:cka-alignment}, joint training produces consistently higher
text-image CKA than generation-only training. This result indicates that
understanding supervision improves generation by better aligning the generation representations with language semantics.

\subsection{A Third Design: Task-decoupled \mot{}}

The opposite failure modes motivate a third design: the task-decoupled \mot{}
illustrated in \paperfigref{fig:architectures}(c). It routes text and clean
image tokens (for understanding or serving as context) through the pretrained LLM branch,
while the noised image tokens (for generation) use a generation-specialized visual branch.
Thus, understanding retains direct access to pretrained language semantics,
whereas generation receives parameters specialized for the generation objective.

The visual representations in the two branches remain softly aligned in two ways. First, they attend to the
same textual features, providing a shared semantic context. Second, with image editing and interleaved samples, the context image in the understanding branch acts as a condition for the generation branch via shared attention at each layer.
To further strengthen the connection between them, we allocate an additional 20\% of the training mixture to an image
reconstruction task formulated as image editing. The target images are drawn
from the text-to-image data, while 50\% of the visual tokens in each
corresponding context image are randomly masked.

As shown in \papertabref{tab:representation-results} (Row 5), the task-decoupled \mot{} avoids the
opposite trade-offs observed in the dense and modality-decoupled models. This result suggests that the interference is not
intrinsic to unification. Instead, it can be mitigated by specializing the
visual computation required by the two objectives while retaining pathways
for semantic interaction. We do not claim that task-decoupled routing is the
unique optimal solution for unified modeling; other conditional architectures, such as
mixtures-of-experts (MoE) \citep{moe} with task- or token-dependent routing, may also achieve a balance between specialization and sharing, which we leave for future work.

\begin{findingbox}
  \findinglabel{Finding 1 (Representation Level):} {Visual understanding and generation can benefit
  each other, but mutual gains depend on the architecture.}
  Selectively specializing conflicting visual pathways while preserving semantic interaction enables both capabilities to improve without either dominating the other.
\end{findingbox}

We therefore use the task-decoupled \mot{} as the default base model for the
task-level and system-level studies that follow.

\section{Task-Level Synergy}
\label{sec:task}

Having obtained a model that avoids conflict between visual understanding and
generation, we next move from representations to tasks and ask: when these two learning objectives rely on the same domain knowledge or underlying capability, can
supervision in one direction improve the other? We study this question through
three cases.

Across all three cases, 50\% of each training mixture consists of the same
general-purpose data used in the representation-level training stage, after
removing examples related to the task or domain under study. The remaining
mixture consists of the corresponding task-specific understanding and
generation data. This preserves the model's general capabilities while
preventing overlapping auxiliary data from confounding the task-level
comparison.

\subsection{Case I: Geometry Problem Solving}

Our first case asks whether the generation objective can improve
geometry reasoning. Geometry problem solving requires the model to identify entities, relations, and constraints in geometric diagrams, while generating
or editing a geometric figure requires it to construct visual content that
satisfies the same structure. The two objectives therefore partially share the same
domain knowledge.

We construct a geometry-related understanding data mixture which contains 965K geometry problem-solving data and general
mathematical-reasoning data selected from public datasets \citep{lin2026mmfinereason,Geometry3K,zhang2025mavis,wiedmann2025finevision,numina_math_datasets,virl39k,honeybee}. The generation part consists of two generation datasets:
text-to-image examples from \textsc{MathCanvas-Imagen} \citep{shi2026mathcanvas} and geometry-diagram
editing examples from \textsc{MathCanvas-Edit} \citep{shi2026mathcanvas}. Our primary comparison is
between understanding-only training (\und{}) and joint training
(\joint{}) to see whether the generation objective can effectively improve the geometry understanding and reasoning abilities.

We then ask a more specific question: with the same data, is the generation objective a more
effective way than the corresponding understanding objective in this case? To test this, we
convert exactly the same generation data into an understanding format.
Text-to-image pairs become image-description tasks, and editing pairs become
tasks that describe the transformation between the source and target images.
This Textual-Desc baseline controls the source data while changing
only how the model learns from it: through text prediction instead of image
generation.

\begin{table}[h]
  \caption{\textbf{Geometry problem solving and general mathematical
  reasoning.} \joint{} adds the geometric image-generation
  objectives. \und{} + Textual-Desc uses the same examples after
  converting them into image-description and edit-description tasks. Higher
  is better for all metrics.}
  \label{tab:geometry-results}
  \centering
  \small
  \setlength{\tabcolsep}{4pt}
  \begin{tabular}{lcccc}
    \toprule
    \multirow{2}{*}{Training} & \multicolumn{2}{c}{Geometry}
      & \multicolumn{2}{c}{General Math} \\
    \cmidrule(lr){2-3}\cmidrule(lr){4-5}
    & Geometry3K & PGPS9K & MathVerse$_{\text{testmini}}$
      & MathVista$_{\text{testmini}}$ \\
    \midrule
    \und{} & 59.90 & 72.40 & 60.13 & 71.80 \\
    \und{} + Textual-Desc
      & 63.56 & {73.90} & 59.11 & 73.60 \\
    \joint{} & {65.39} & {73.40} & {63.15} & {74.30} \\
    \bottomrule
  \end{tabular}
\end{table}

We evaluate the models on geometry problem solving benchmark including Geometry3K \citep{Geometry3K} and PGPS9K \citep{PGPS9K}, and general math benchmarks including MathVerse \citep{zhang2024mathverse} and MathVista \citep{lu2024mathvista}. As shown in \papertabref{tab:geometry-results}, adding the generation objective
improves the understanding model on all four benchmarks. The gains also transfer beyond
the directly paired geometry tasks into general math reasoning. Converting the same additional examples into understanding tasks produces a
smaller and less consistent benefit. This comparison shows that the benefit is not solely a result
of exposing the model to more geometry data. For these tasks and data,
learning to generate and edit the diagrams is a more effective form of
auxiliary supervision for geometry and general mathematical reasoning than
learning text-output descriptions.

\subsection{Case II: SVG Understanding and Generation}

Scalable Vector Graphics (SVG) provides a naturally bidirectional domain for studying task-level
transfer: the same icon can be represented either as a rendered image or as an
executable vector program. We ask whether learning both mappings jointly
improves the model's correspondence between symbolic SVG structure and visual
appearance.

\paragraph{Task setup.}
In the understanding direction, the model receives a rendered icon image and
outputs its SVG code. In the generation direction,
the model receives SVG code and directly renders the corresponding image. Although the
output spaces differ, both directions depend on the same latent scene graph
and rendering relations. We use the 359K SVG code-image pairs from UniSVG \citep{li2025unisvg} to construct the corresponding understanding and generation training data. For evaluation, we use the test data of UniSVG and adopt its metric to evaluate the directly generated or code-rendered icon, which is a weighted sum of SSIM \citep{SSIM}, LPIPS \citep{LPIPS}, and CLIP similarity \citep{clip2021}.

\begin{wraptable}{r}{0.48\linewidth}
  \centering
  \vspace{-0.7\baselineskip}
  \caption{SVG understanding and generation results. Higher is better.}
  \label{tab:svg-results}
  \small
  \setlength{\tabcolsep}{4pt}
  \begin{tabular}{lcc}
    \toprule
    Training & Image$\rightarrow$SVG & SVG$\rightarrow$Image \\
    \midrule
    \und{}/\gen{} & 80.64 & 80.21 \\
    \joint{} & 81.70 & 86.52 \\
    \bottomrule
  \end{tabular}
  \vspace{-0.6\baselineskip}
\end{wraptable}

\paragraph{Bidirectional transfer.}
We compare understanding-only, generation-only, and joint training in
\papertabref{tab:svg-results}. Joint training improves both SVG image understanding
and SVG-conditioned image generation over their corresponding
single-direction models. This mutual improvement suggests that the two tasks
share more than generic image--text semantics: supervision in either direction
helps the model learn the executable relationship between vector operations
and their visual consequences.

\paragraph{Diagnostic: mental visualization from SVG code.}
We hypothesize that joint training improves the model's ability to mentally
visualize the corresponding image directly from SVG code. We test this hypothesis
with two complementary diagnostics. First, we construct a code-conditioned
multiple-choice blind VQA benchmark in which each question asks about the appearance
implied by an SVG program without providing its rendered image. Crucially, the
answers cannot be obtained through a surface-level reading of tags or
individual attributes. The model must mentally execute and compose operations
such as path rendering, geometric transformations, grouping, and layering to
infer the final visual outcome. Details and examples about this benchmark are provided in Section \ref{app:tasks_details_svg}. Second, we inspect the predicted clean image
after only the first generation step. This early prediction indicates how
readily the model forms a global visual plan directly from the code.

\begin{figure*}[t]
  \centering
  \begin{minipage}[t]{0.35\textwidth}
    \centering
    \vspace{0pt}
    \includegraphics[width=\linewidth]{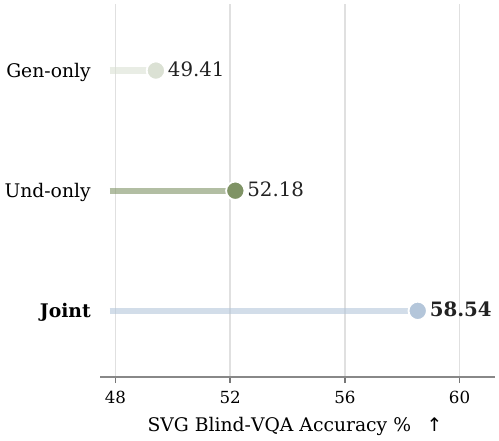}
  \end{minipage}
  \hfill
  \begin{minipage}[t]{0.62\textwidth}
    \centering
    \vspace{0pt}
    \includegraphics[width=\linewidth]{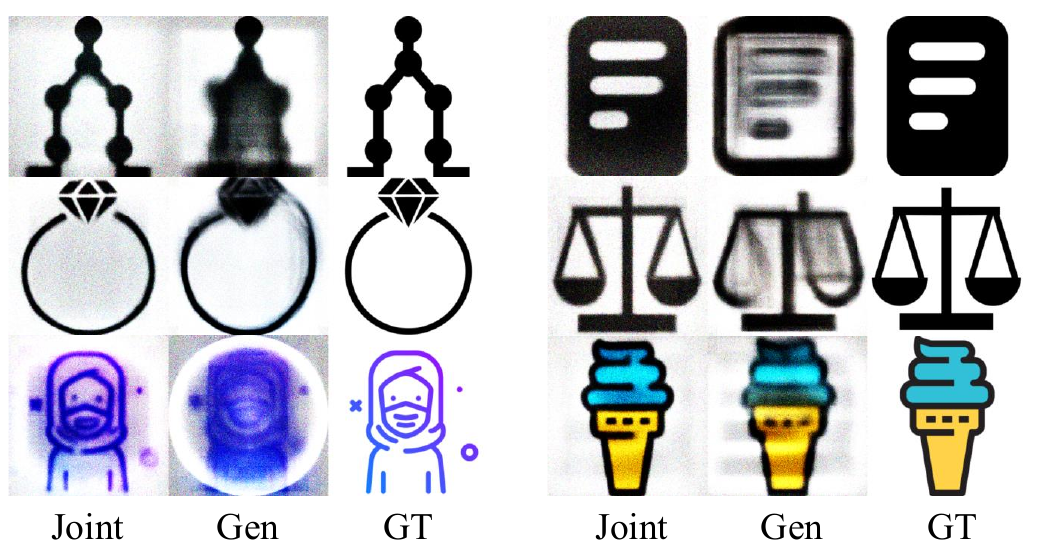}
  \end{minipage}
  \caption{\textbf{Joint training improves direct visualization from SVG
  code.} Left: Joint training achieves higher accuracy on code-conditioned VQA,
  which requires inferring the rendered appearance of an SVG program without
  seeing the image. Right: The predicted clean image after only one generation step
  visualizes how easily the model forms global shape and layout from the same
  code.}
  \label{fig:svg-mental-visualization}
\end{figure*}

As shown in \paperfigref{fig:svg-mental-visualization}, the \joint{} model
performs better on the blind VQA diagnostic and recovers recognizable contours and
spatial layout earlier than \gen{}. Together, these results provide
complementary evidence that joint training strengthens an internal, visually
grounded simulation of SVG execution: the model can both reason about the
implied image in text and instantiate its structure earlier during generation.

\subsection{Case III: 3D spatial intelligence}

For our third case study, we move beyond the specialized domains of geometry
problems and SVGs to 3D spatial intelligence, a broader and more widely required
capability that is fundamental to many applications like embodied agents and interactive world modeling. We examine whether understanding and
generation can improve each other on general 3D spatial intelligence tasks. Although the tasks are not exact input--output inverses, they
partially share the underlying capability to perceive and understand the 3D space.

\paragraph{Task setup.}
The understanding training mixture for 3D SI contains single-image, multi-image, and video-based
3D VQA, covering spatial relations, viewpoint reasoning, depth, and camera
motion. 
For the generation mixture, we design the following four generative tasks related to 3D SI. Examples of these tasks are provided in Section \ref{app:tasks_details_si}.
\begin{itemize}
  \item \textbf{Ego-motion Transition}, which requires the model to predict the target frame based on a source frame and ego movement descriptions. We use the public dataset SpatialEdit-500K \citep{xiao2026spatialedit} and additionally construct 1.6M samples from Scannet \citep{scannet} and Scannet++ \citep{scannet++}.
  \item \textbf{Multi-view Reconstruction}, which requires the model to predict a target view based on the provided views of a certain object. We use the Objaverse \citep{objaverse} data to create 248K samples. We only consider front, back, left, and right views and randomly choose 2 views as context and 1 view as target.
  \item \textbf{View Sequence Completion (VSC)}, which requires the model to predict the middle frame given the first and last 2 frames in a 5-frame sequence. We construct 880K samples from Scannet \citep{scannet}, Scannet++ \citep{scannet++}, and ARKitScenes \citep{dehghan2021arkitscenes}.
  \item \textbf{Layout-to-image Generation}, which requires the model to
  generate an image conditioned on a description of its 3D layout. We consider
  two forms of layout conditioning: (i) a structured representation specifying
  the size of every object and its 3D position relative to the camera, and
  (ii) a natural-language description emphasizing the scene's 3D structure
  and spatial relations. Using Omni3D \citep{brazil2023omni3d}, we construct
  332K training samples across these two formats.
\end{itemize}

We evaluate 3D spatial understanding on nine benchmarks including VSI-Bench \citep{VSI}, SPAR-Bench \citep{spar}, CV-Bench \citep{tong2024cambrian}, DA-2K \citep{DA2K}, MindCube \citep{mindcube}, ViewSpatial \citep{viewspatial}, 3DSRBench \citep{ma20253dsrbench}, MMSI \citep{yang2025mmsi}, All-Angles \citep{allangles}. For generation, we evaluate ego-motion transition on
SpatialEdit benchmark using Viewpoint Error (VE) and Framing Error (FE). We also construct a test set of the VSC task and evaluate using PSNR, SSIM, and LPIPS as metrics.

\begin{table}[h]
  \caption{{Understanding results on 3D SI
  benchmarks.} Higher is better for all benchmarks.}
  \label{tab:3d-understanding}
  \centering
  \small
  \setlength{\tabcolsep}{3.2pt}
  \renewcommand{\arraystretch}{1.08}
  \begin{tabular}{lcccccccccc}
    \toprule
    Model
      & \rotatebox[origin=c]{60}{VSI-Bench}
      & \rotatebox[origin=c]{60}{SPAR-Bench}
      & \rotatebox[origin=c]{60}{CV-Bench}
      & \rotatebox[origin=c]{60}{DA-2K}
      & \rotatebox[origin=c]{60}{MindCube-tiny}
      & \rotatebox[origin=c]{60}{ViewSpatial}
      & \rotatebox[origin=c]{60}{3DSRBench}
      & \rotatebox[origin=c]{60}{MMSI}
      & \rotatebox[origin=c]{60}{ALL-Angles}
      & \rotatebox[origin=c]{60}{\textbf{Average}} \\
    \midrule
    \und{} only
      & 53.98 & 38.80 & 77.67 & 70.69 & 75.71
      & 58.75 & 59.84 & 36.40 & 42.50 & \textbf{57.15} \\
    \joint{}
      & 55.49 & 41.13 & 78.77 & 73.89 & 77.90
      & 58.07 & 60.51 & 39.20 & 46.11 & \textbf{59.01} \\
    \bottomrule
  \end{tabular}
\end{table}

\begin{table}[t]
  \caption{Benchmark results on 3D SI related generation tasks.}
  \label{tab:3d-generation}
  \centering
  \setlength{\tabcolsep}{6pt}
  \begin{tabular}{lccccc}
    \toprule
    \multirow{2}{*}{Model}
      & \multicolumn{2}{c}{Spatial-Edit}
      & \multicolumn{3}{c}{View Sequence Completion} \\
    \cmidrule(lr){2-3}\cmidrule(lr){4-6}
    & VE $\downarrow$ & FE $\downarrow$
      & PSNR $\uparrow$ & SSIM $\uparrow$ & LPIPS $\downarrow$ \\
    \midrule
    \gen{} only
      & 0.4062 & 0.6564 & 16.76 & 0.6365 & 0.3828 \\
    \joint{}
      & 0.3766 & 0.6297 & 19.15 & 0.6674 & 0.3078 \\
    \bottomrule
  \end{tabular}
\end{table}

\paragraph{Bidirectional transfer.}
For understanding
(\papertabref{tab:3d-understanding}), joint training improves eight of nine benchmarks and
raises the average score from 57.15 to 59.01. For generation
(\papertabref{tab:3d-generation}), it reduces both errors on Spatial-Edit and
improves all three VSC metrics. These results show that 3D SI-related visual understanding and generation tasks benefit each other as the underlying knowledge overlaps.

\begin{findingbox}
  \findinglabel{Finding 2 (Task Level):} Shared underlying knowledge or capabilities make understanding and generation tasks mutually reinforcing.
\end{findingbox}

This finding motivates exploring unified learning in broader settings where
multiple capabilities depend on shared latent knowledge or structure. One
direct example is world modeling and policy/action learning, where
understanding the current state, predicting future observations, and
generating actions all rely on a common model of the physical world and its
dynamics.

\section{System-Level Synergy}
\label{sec:system}

We now move beyond tasks centered primarily on either understanding or
generation to a higher-level setting in which both capabilities are inherently
required to operate sequentially and interactively. Some of these tasks do not necessarily require a unified model: separate
understanding and generation models can be composed into an agentic pipeline. Our system-level question is therefore whether, under a fair and
matched comparison, an end-to-end UMM provides a measurable performance
advantage over composing specialized understanding and generation models,
beyond the practical simplicity of using a single model.

We study this question through the reasoning-intensive image editing task. Given a
source image and an implicit editing request, the system must understand the source,
reason about the intended concrete edit, and generate the target image. This task
admits both an end-to-end solution and a natural planner--executor
decomposition, making it suitable for comparing unified and agentic systems.

\subsection{Controlled comparison setup}

To compare end-to-end unification and modular composition under controlled
conditions, we initialize all models from the same task-decoupled \mot{} checkpoint
and transform the same set of reasoning-editing examples into three training
formats:
\begin{enumerate}
  \item \textbf{End-to-end understanding and generation:} given a source image
  and the original implicit instruction, the model reasons about the requested
  change, produces an explicit edit instruction, and then generates the target
  image.
  \item \textbf{Understanding/planning:} given the same source image and
  implicit instruction, the model performs the reasoning and outputs only the
  explicit edit instruction.
  \item \textbf{Generation/execution:} given the source image, implicit
  instruction, and explicit edit instruction, the model generates the target
  image.
\end{enumerate}
We train one model for each format. At inference time, the agentic pipeline
composes the planner in (2) with the executor in (3), while the model in (1)
performs the complete process end to end. In addition to the reasoning-editing
examples, 50\% of each training mixture consists of the same general-purpose
data used in the representation-level training stage.

\begin{table*}[t]
  \caption{Results on reasoning-intensive image editing benchmarks. Higher is better for all results.}
  \label{tab:system-results}
  \centering
  \scriptsize
  \setlength{\tabcolsep}{4.5pt}
  \renewcommand{\arraystretch}{1.08}
  \begin{tabular}{lccccccccc}
    \toprule
    \multirow{2}{*}{System}
      & \multicolumn{5}{c}{RISEBench}
      & \multicolumn{4}{c}{KRIS-Bench} \\
    \cmidrule(lr){2-6}\cmidrule(lr){7-10}
      & Temporal & Causal & Spatial & Logical & \textbf{Overall}
      & Factual & Conceptual & Procedural & \textbf{Overall} \\
    \midrule
    Planner $\rightarrow$ Executor
      & 18.82 & 22.22 & 14.00 & 11.76 & \textbf{16.66}
      & 61.40 & 71.17 & 64.58 & \textbf{66.48} \\
    End-to-end \umm{}
      & 22.35 & 27.77 & 16.00 & 9.41 & \textbf{18.88}
      & 64.73 & 72.32 & 66.16 & \textbf{68.33} \\
    \bottomrule
  \end{tabular}
\end{table*}

We evaluate both systems on RISEBench \citep{risebench2025} and KRIS-Bench
\citep{krisbench2025}. As shown in \papertabref{tab:system-results}, the
end-to-end \umm{} outperforms the planner--executor pipeline on both
benchmarks. Under the matched base model and task data, this result shows that
end-to-end unification can provide a performance advantage when visual
understanding, reasoning, and generation must interact closely within a task.

\begin{findingbox}
  \findinglabel{Finding 3 (System Level):} End-to-end unification can outperform
  modular composition for complex tasks that explicitly require both visual understanding and generation.
\end{findingbox}

A natural future direction is to internalize broader agentic generation processes, such
as multimodal search + generation and iterative visual refinement, as native
capabilities of UMMs. This could enable increasingly complex tasks to be
learned and completed end-to-end within a unified model.

\section{Related Work}
\label{sec:related}

\paragraph{Unified multimodal models.}
Unified multimodal models (UMMs) integrate visual understanding and generation
within a single model and have developed rapidly in recent years \citep{chameleon2024,showo2024,janus2024,transfusion2024,bagel2025,liu2025tuna,sensenova2026sensenovau1}. By understanding and generating text
and images within a common context, these models enable distinctive
capabilities such as generating interleaved multimodal documents and combining
textual reasoning with visual reasoning performed through generation. However,
such functional unification does not establish whether the underlying visual
understanding and generation capabilities genuinely reinforce each other or
merely coexist in the same model. We study when and why joint learning produces
synergy, and when architectural sharing instead leads to interference.

\paragraph{Native multimodal models.}
The term \emph{native} can describe both how a multimodal model is trained and how it is
structured. From a training perspective, native models are exposed to
multimodal data from the beginning of pretraining, rather than first
pretraining a text-only LLM and introducing multimodal capabilities
afterward \citep{kimi3,tong2026beyond,thinkingmachines2026interactionmodels,han2026towards}. From a structural
perspective, visual inputs and outputs are processed directly by the multimodal
backbone, reducing reliance on external encoders, image tokenizers, or VAEs.
Recent systems explore native pixel representations, unified visual features,
and increasingly end-to-end multimodal training
\citep{fuyu-8b,thinkingmachines2026interactionmodels,neoov2026,sensenova2026sensenovau1}.
We adopt a structurally native, pixel-in, pixel-out setting without a
pretrained visual encoder or generative VAE, allowing the interaction between
understanding and generation to be studied without external visual priors. We
do not, however, train the entire model natively from scratch. Pretraining a
language model with sufficiently strong language and reasoning capabilities
requires substantial data and computation, while our controlled analyses and
downstream tasks depend on such capabilities to yield meaningful results. We
therefore initialize from a pretrained LLM and focus specifically on how
visual understanding and generation are learned and interact.

\paragraph{Understanding and generation: conflict or synergy?}
Existing work provides no clear consensus on whether visual understanding and
generation intrinsically benefit each other. Some studies report substantial
interference between the two objectives and motivate explicitly decoupling
their representations or computation pathways \citep{janus2024,metaquery}.
Others introduce additional architectural components or training objectives to
transfer understanding signals to generation. For example, Reconstruction
Alignment (RecA) conditions image reconstruction on dense understanding
features \citep{RecA}, while UNO injects captioning and visual-regression
supervision into generative representations \citep{uno}. Concurrent
studies further demonstrate that the two capabilities can mutually improve through knowledge flow,
but primarily under carefully designed tasks \citep{kang2026transferability,han2026towards}.

Together, these findings suggest that synergy is possible but highly
conditional. However, the observed effects are often entangled with
specialized modules, auxiliary losses, task-specific formulations, or changes
in pretrained visual components, making the intrinsic relationship between
understanding and generation difficult to isolate. We address this gap in a
controlled native setting, studying their interaction at
the representation, task, and system levels.

\section{Limitations}
\label{sec:limitations}

Our study focuses on the currently prevalent modeling formulation
that combines discrete autoregressive modeling for text with continuous
diffusion-based modeling for images. Although most of our conclusions are
likely to transfer to other formulations, such as fully discrete or fully
autoregressive multimodal modeling, their generality remains to be empirically
validated. In addition, our experiments mainly use a task-decoupled architecture to
demonstrate that understanding and generation can mutually benefit. We do not extensively explore the broader architecture
design space; identifying the optimal unified architecture and balance between
sharing and specialization remains an important direction for future work.

\section{Conclusion}

We studied understanding--generation synergy in native UMMs across
representations, tasks, and systems. We find that their mutual benefit is
architecture-dependent, and task-aware specialization enables both
capabilities to coexist without asymmetric degradation. Joint training further
improves both directions when tasks share underlying knowledge, while an
end-to-end UMM outperforms a matched modular pipeline when understanding,
reasoning, and generation must interact closely. These results demonstrate
that UMMs offer value beyond a unified interface by enabling knowledge transfer
and end-to-end optimization across the two capabilities.

\bibliography{iclr2027_conference}

@article{bagel2025,
  title={Emerging properties in unified multimodal pretraining},
  author={Deng, Chaorui and Zhu, Deyao and Li, Kunchang and Gou, Chenhui and Li, Feng and Wang, Zeyu and Zhong, Shu and Yu, Weihao and Nie, Xiaonan and Song, Ziang and others},
  journal={arXiv preprint arXiv:2505.14683},
  year={2025}
}

@article{sensenova2026sensenovau1,
  title        = {SenseNova-U1: Unifying Multimodal Understanding and Generation with NEO-unify Architecture},
  author       = {Diao, Haiwen and Wu, Penghao and Deng, Hanming and Wang, Jiahao and Bai, Shihao and Wu, Silei and Fan, Weichen and Ye, Wenjie and Tong, Wenwen and Fan, Xiangyu and others},
  journal      = {arXiv preprint arXiv:2605.12500},
  year         = {2026}
}

@article{cui2025emu35,
  title={Emu3. 5: Native multimodal models are world learners},
  author={Cui, Yufeng and Chen, Honghao and Deng, Haoge and Huang, Xu and Li, Xinghang and Liu, Jirong and Liu, Yang and Luo, Zhuoyan and Wang, Jinsheng and Wang, Wenxuan and others},
  journal={arXiv preprint arXiv:2510.26583},
  year={2025}
}

@article{xin2025lumina,
  title={Lumina-dimoo: An omni diffusion large language model for multi-modal generation and understanding},
  author={Xin, Yi and Qin, Qi and Luo, Siqi and Zhu, Kaiwen and Yan, Juncheng and Tai, Yan and Lei, Jiayi and Cao, Yuewen and Wang, Keqi and Wang, Yibin and others},
  journal={arXiv preprint arXiv:2510.06308},
  year={2025}
}

@article{liu2025tuna,
  title={Tuna: Taming unified visual representations for native unified multimodal models},
  author={Liu, Zhiheng and Ren, Weiming and Liu, Haozhe and Zhou, Zijian and Chen, Shoufa and Qiu, Haonan and Huang, Xiaoke and An, Zhaochong and Yang, Fanny and Patel, Aditya and others},
  journal={arXiv preprint arXiv:2512.02014},
  year={2025}
}

@article{chameleon2024,
  title={Chameleon: Mixed-modal early-fusion foundation models},
  author={Team, Chameleon},
  journal={arXiv preprint arXiv:2405.09818},
  year={2024}
}

@inproceedings{Harmon,
  title={Harmonizing visual representations for unified multimodal understanding and generation},
  author={Wu, Size and Zhang, Wenwei and Xu, Lumin and Jin, Sheng and Wu, Zhonghua and Tao, Qingyi and Liu, Wentao and Li, Wei and Loy, Chen Change},
  booktitle={ICCV},
  year={2025}
}

@article{ai2026llada2,
  title={Llada2. 0-uni: Unifying multimodal understanding and generation with diffusion large language model},
  author={AI, Inclusion and Bie, Tiwei and Chen, Haoxing and Chen, Tieyuan and Cheng, Zhenglin and Cui, Long and Gan, Kai and Huang, Zhicheng and Lan, Zhenzhong and Li, Haoquan and others},
  journal={arXiv preprint arXiv:2604.20796},
  year={2026}
}

@article{mingunivision,
  title={Ming-univision: Joint image understanding and generation with a unified continuous tokenizer},
  author={Huang, Ziyuan and Zheng, DanDan and Zou, Cheng and Liu, Rui and Wang, Xiaolong and Ji, Kaixiang and Chai, Weilong and Sun, Jianxin and Wang, Libin and Lv, Yongjie and others},
  journal={arXiv preprint arXiv:2510.06590},
  year={2025}
}

@inproceedings{cka2019,
  title={Similarity of neural network representations revisited},
  author={Kornblith, Simon and Norouzi, Mohammad and Lee, Honglak and Hinton, Geoffrey},
  booktitle={ICML},
  year={2019}
}

@inproceedings{clip2021,
  title={Learning transferable visual models from natural language supervision},
  author={Radford, Alec and Kim, Jong Wook and Hallacy, Chris and Ramesh, Aditya and Goh, Gabriel and Agarwal, Sandhini and Sastry, Girish and Askell, Amanda and Mishkin, Pamela and Clark, Jack and others},
  booktitle={ICML},
  year={2021}
}

@inproceedings{janus2024,
  title={Janus: Decoupling visual encoding for unified multimodal understanding and generation},
  author={Wu, Chengyue and Chen, Xiaokang and Wu, Zhiyu and Ma, Yiyang and Liu, Xingchao and Pan, Zizheng and Liu, Wen and Xie, Zhenda and Yu, Xingkai and Ruan, Chong and others},
  booktitle={CVPR},
  year={2025}
}

@article{krisbench2025,
  title={Kris-bench: Benchmarking next-level intelligent image editing models},
  author={Wu, Yongliang and Li, Zonghui and Hu, Xinting and Ye, Xinyu and Zeng, Xianfang and Yu, Gang and Zhu, Wenbo and Schiele, Bernt and Yang, Ming-Hsuan and Yang, Xu},
  journal={NeurIPS},
  year={2026}
}

@misc{discus0434_aesthetic_predictor_v2_5,
  author       = {discus0434},
  title        = {Aesthetic Predictor V2.5},
  howpublished = {\url{https://github.com/discus0434/aesthetic-predictor-v2-5}},
  year         = {2024},
  note         = {SigLIP-based Aesthetic Score Predictor}
}

@INPROCEEDINGS{ImageNet,
  author={Deng, Jia and Dong, Wei and Socher, Richard and Li, Li-Jia and Kai Li and Li Fei-Fei},
  booktitle={CVPR}, 
  title={ImageNet: A large-scale hierarchical image database}, 
  year={2009}}

@inproceedings{ade20k,
  title={Scene parsing through ade20k dataset},
  author={Zhou, Bolei and Zhao, Hang and Puig, Xavier and Fidler, Sanja and Barriuso, Adela and Torralba, Antonio},
  booktitle={CVPR},
  year={2017},
}

@inproceedings{nyudepthv2,
  title={Indoor segmentation and support inference from rgbd images},
  author={Silberman, Nathan and Hoiem, Derek and Kohli, Pushmeet and Fergus, Rob},
  booktitle={ECCV},
  year={2012},
}

@article{januspro,
  title={Janus-pro: Unified multimodal understanding and generation with data and model scaling},
  author={Chen, Xiaokang and Wu, Zhiyu and Liu, Xingchao and Pan, Zizheng and Liu, Wen and Xie, Zhenda and Yu, Xingkai and Ruan, Chong},
  journal={arXiv preprint arXiv:2501.17811},
  year={2025}
}

@article{moe,
  title={Outrageously large neural networks: The sparsely-gated mixture-of-experts layer},
  author={Shazeer, Noam and Mirhoseini, Azalia and Maziarz, Krzysztof and Davis, Andy and Le, Quoc and Hinton, Geoffrey and Dean, Jeff},
  journal={arXiv preprint arXiv:1701.06538},
  year={2017}
}

@article{geneval2,
  title={Geneval 2: Addressing benchmark drift in text-to-image evaluation},
  author={Kamath, Amita and Chang, Kai-Wei and Krishna, Ranjay and Zettlemoyer, Luke and Hu, Yushi and Ghazvininejad, Marjan},
  journal={arXiv preprint arXiv:2512.16853},
  year={2025}
}

@article{mot2025,
  title={Mixture-of-transformers: A sparse and scalable architecture for multi-modal foundation models},
  author={Liang, Weixin and Yu, Lili and Luo, Liang and Iyer, Srini and Dong, Ning and Zhou, Chunting and Ghosh, Gargi and Lewis, Mike and Yih, Wen-tau and Zettlemoyer, Luke and others},
  journal={Transactions on Machine Learning Research},
  year={2024}
}

@inproceedings{neo,
  title={From pixels to words--towards native vision-language primitives at scale},
  author={Diao, Haiwen and Li, Mingxuan and Wu, Silei and Dai, Linjun and Wang, Xiaohua and Deng, Hanming and Lu, Lewei and Lin, Dahua and Liu, Ziwei},
  booktitle={ICLR},
  year={2026}
}

@article{neoov2026,
  title={From Pixels to Words--Towards Native One-Vision Models at Scale},
  author={Diao, Haiwen and Wang, Jiahao and Wu, Penghao and Dong, Yuhao and Niu, Yuwei and Zhu, Yue and Cai, Zhongang and Fan, Weichen and Dai, Linjun and Wu, Silei and others},
  journal={arXiv preprint arXiv:2605.28820},
  year={2026}
}

@article{RecA,
  title={Reconstruction Alignment Improves Unified Multimodal Models},
  author={Xie, Ji and Darrell, Trevor and Zettlemoyer, Luke and Wang, XuDong},
  journal={arXiv preprint arXiv:2509.07295},
  year={2025}
}

@article{metaquery,
  title={Transfer between modalities with metaqueries},
  author={Pan, Xichen and Shukla, Satya Narayan and Singh, Aashu and Zhao, Zhuokai and Mishra, Shlok Kumar and Wang, Jialiang and Xu, Zhiyang and Chen, Jiuhai and Li, Kunpeng and Juefei-Xu, Felix and others},
  journal={arXiv preprint arXiv:2504.06256},
  year={2025}
}

@article{blip3o,
  title={Blip3-o: A family of fully open unified multimodal models-architecture, training and dataset},
  author={Chen, Jiuhai and Xu, Zhiyang and Pan, Xichen and Hu, Yushi and Qin, Can and Goldstein, Tom and Huang, Lifu and Zhou, Tianyi and Xie, Saining and Savarese, Silvio and others},
  journal={arXiv preprint arXiv:2505.09568},
  year={2025}
}

@article{qwen3_2025,
  title={Qwen3 technical report},
  author={Yang, An and Li, Anfeng and Yang, Baosong and Zhang, Beichen and Hui, Binyuan and Zheng, Bo and Yu, Bowen and Gao, Chang and Huang, Chengen and Lv, Chenxu and others},
  journal={arXiv preprint arXiv:2505.09388},
  year={2025}
}

@article{risebench2025,
  title={Envisioning beyond the pixels: Benchmarking reasoning-informed visual editing},
  author={Zhao, Xiangyu and Zhang, Peiyuan and Tang, Kexian and Zhu, Xiaorong and Li, Hao and Chai, Wenhao and Zhang, Zicheng and Xia, Renqiu and Zhai, Guangtao and Yan, Junchi and others},
  journal={NeurIPS},
  year={2026}
}

@inproceedings{showo2024,
  title={Show-o: One single transformer to unify multimodal understanding and generation},
  author={Xie, Jinheng and Mao, Weijia and Bai, Zechen and Zhang, David Junhao and Wang, Weihao and Lin, Kevin Qinghong and Gu, Yuchao and Chen, Zhijie and Yang, Zhenheng and Shou, Mike Zheng},
  booktitle={ICLR},
  year={2025}
}

@inproceedings{transfusion2024,
  title={Transfusion: Predict the next token and diffuse images with one multi-modal model},
  author={Zhou, Chunting and Yu, Lili and Babu, Arun and Tirumala, Kushal and Yasunaga, Michihiro and Shamis, Leonid and Kahn, Jacob and Ma, Xuezhe and Zettlemoyer, Luke and Levy, Omer},
  booktitle={ICLR},
  year={2025}
}

@article{DPG,
  title={Ella: Equip diffusion models with llm for enhanced semantic alignment},
  author={Hu, Xiwei and Wang, Rui and Fang, Yixiao and Fu, Bin and Cheng, Pei and Yu, Gang},
  journal={arXiv preprint arXiv:2403.05135},
  year={2024}
}

@inproceedings{ma2025hpsv3,
  title={Hpsv3: Towards wide-spectrum human preference score},
  author={Ma, Yuhang and Wu, Xiaoshi and Sun, Keqiang and Li, Hongsheng},
  booktitle={ICCV},
  year={2025},
}

@article{xiao2026spatialedit,
  title={Spatialedit: Benchmarking fine-grained image spatial editing},
  author={Xiao, Yicheng and Zhang, Wenhu and Song, Lin and Chen, Yukang and Li, Wenbo and Jiang, Nan and Ren, Tianhe and Lin, Haokun and Huang, Wei and Huang, Haoyang and others},
  journal={arXiv preprint arXiv:2604.04911},
  year={2026}
}

@article{p2gb,
  title={Plug-and-play grounding of reasoning in multimodal large language models},
  author={Chen, Jiaxing and Liu, Yuxuan and Li, Dehu and An, Xiang and Deng, Weimo and Feng, Ziyong and Zhao, Yongle and Xie, Yin},
  journal={arXiv preprint arXiv:2403.19322},
  year={2024}
}

@inproceedings{mmmu,
  title={Mmmu: A massive multi-discipline multimodal understanding and reasoning benchmark for expert agi},
  author={Yue, Xiang and Ni, Yuansheng and Zhang, Kai and Zheng, Tianyu and Liu, Ruoqi and Zhang, Ge and Stevens, Samuel and Jiang, Dongfu and Ren, Weiming and Sun, Yuxuan and others},
  booktitle={CVPR},
  year={2024}
}

@inproceedings{ai2d,
  title={A diagram is worth a dozen images},
  author={Kembhavi, Aniruddha and Salvato, Mike and Kolve, Eric and Seo, Minjoon and Hajishirzi, Hannaneh and Farhadi, Ali},
  booktitle={ECCV},
  year={2016},
}

@article{seedbench,
  title={Seed-bench: Benchmarking multimodal llms with generative comprehension},
  author={Li, Bohao and Wang, Rui and Wang, Guangzhi and Ge, Yuying and Ge, Yixiao and Shan, Ying},
  journal={arXiv preprint arXiv:2307.16125},
  year={2023}
}

@inproceedings{docvqa,
  title={Docvqa: A dataset for vqa on document images},
  author={Mathew, Minesh and Karatzas, Dimosthenis and Jawahar, CV},
  booktitle={WACV},
  year={2021}
}

@inproceedings{chartqa,
    title = "{C}hart{QA}: A Benchmark for Question Answering about Charts with Visual and Logical Reasoning",
    author = "Masry, Ahmed  and
      Long, Do  and
      Tan, Jia Qing  and
      Joty, Shafiq  and
      Hoque, Enamul",
    booktitle = "ACL Findings",
    year = "2022",
}

@inproceedings{infovqa,
  title={Infographicvqa},
  author={Mathew, Minesh and Bagal, Viraj and Tito, Rub{\`e}n and Karatzas, Dimosthenis and Valveny, Ernest and Jawahar, CV},
  booktitle={WACV},
  year={2022}
}

@article{ocrbench,
    title={OCRBench: on the hidden mystery of OCR in large multimodal models},
    journal={Science China Information Sciences},
    author={Liu, Yuliang and Li, Zhang and Huang, Mingxin and Yang, Biao and Yu, Wenwen and Li, Chunyuan and Yin, Xu-Cheng and Liu, Cheng-Lin and Jin, Lianwen and Bai, Xiang},
    year={2024}}

@article{mme,
  title={MME: A Comprehensive Evaluation Benchmark for Multimodal Large Language Models},
  author={Fu, Chaoyou and Chen, Peixian and Shen, Yunhang and Qin, Yulei and Zhang, Mengdan and Lin, Xu and Yang, Jinrui and Zheng, Xiawu and Li, Ke and Sun, Xing and others},
  journal={arXiv preprint arXiv:2306.13394},
  year={2023}
}

@inproceedings{coco,
  title={Microsoft coco: Common objects in context},
  author={Lin, Tsung-Yi and Maire, Michael and Belongie, Serge and Hays, James and Perona, Pietro and Ramanan, Deva and Doll{\'a}r, Piotr and Zitnick, C Lawrence},
  booktitle={ECCV},
  year={2014},
}

@inproceedings{mmstar,
  title={Are we on the right way for evaluating large vision-language models?},
  author={Chen, Lin and Li, Jinsong and Dong, Xiaoyi and Zhang, Pan and Zang, Yuhang and Chen, Zehui and Duan, Haodong and Wang, Jiaqi and Qiao, Yu and Lin, Dahua and others},
  booktitle={NeurIPS},
  year={2024}
}

@inproceedings{mme-realworld,
  title={Mme-realworld: Could your multimodal llm challenge high-resolution real-world scenarios that are difficult for humans?},
  author={Zhang, Yi-Fan and Zhang, Huanyu and Tian, Haochen and Fu, Chaoyou and Zhang, Shuangqing and Wu, Junfei and Li, Feng and Wang, Kun and Wen, Qingsong and Zhang, Zhang and others},
  booktitle={ICLR},
  year={2025}
}

@inproceedings{MMBench,
  title={Mmbench: Is your multi-modal model an all-around player?},
  author={Liu, Yuan and Duan, Haodong and Zhang, Yuanhan and Li, Bo and Zhang, Songyang and Zhao, Wangbo and Yuan, Yike and Wang, Jiaqi and He, Conghui and Liu, Ziwei and others},
  booktitle={ECCV},
  year={2024},
}

@inproceedings{scannet,
  title={Scannet: Richly-annotated 3d reconstructions of indoor scenes},
  author={Dai, Angela and Chang, Angel X and Savva, Manolis and Halber, Maciej and Funkhouser, Thomas and Nie{\ss}ner, Matthias},
  booktitle={CVPR},
  year={2017}
}

@inproceedings{scannet++,
  title={Scannet++: A high-fidelity dataset of 3d indoor scenes},
  author={Yeshwanth, Chandan and Liu, Yueh-Cheng and Nie{\ss}ner, Matthias and Dai, Angela},
  booktitle={ICCV},
  year={2023}
}

@inproceedings{
dehghan2021arkitscenes,
title={{ARK}itScenes - A Diverse Real-World Dataset for 3D Indoor Scene Understanding Using Mobile {RGB}-D Data},
author={Gilad Baruch and Zhuoyuan Chen and Afshin Dehghan and Tal Dimry and Yuri Feigin and Peter Fu and Thomas Gebauer and Brandon Joffe and Daniel Kurz and Arik Schwartz and Elad Shulman},
booktitle={NeurIPS Datasets and Benchmarks Track},
year={2021},
}

@article{objaverse,
  title={Objaverse: A Universe of Annotated 3D Objects},
  author={Matt Deitke and Dustin Schwenk and Jordi Salvador and Luca Weihs and
          Oscar Michel and Eli VanderBilt and Ludwig Schmidt and
          Kiana Ehsani and Aniruddha Kembhavi and Ali Farhadi},
  journal={arXiv preprint arXiv:2212.08051},
  year={2022}
}

@inproceedings{brazil2023omni3d,
  title={Omni3d: A large benchmark and model for 3d object detection in the wild},
  author={Brazil, Garrick and Kumar, Abhinav and Straub, Julian and Ravi, Nikhila and Johnson, Justin and Gkioxari, Georgia},
  booktitle={CVPR},
  year={2023}
}

@inproceedings{li2025unisvg,
  title={Unisvg: A unified dataset for vector graphic understanding and generation with multimodal large language models},
  author={Li, Jinke and Yu, Jiarui and Wei, Chenxing and Dong, Hande and Lin, Qiang and Yang, Liangjing and Wang, Zhicai and Hao, Yanbin},
  booktitle={Proceedings of the 33rd ACM International Conference on Multimedia},
  year={2025}
}

@inproceedings{shi2026mathcanvas,
  title={Mathcanvas: Intrinsic visual chain-of-thought for multimodal mathematical reasoning},
  author={Shi, Weikang and Yu, Aldrich and Fang, Rongyao and Ren, Houxing and Wang, Ke and Zhou, Aojun and Tian, Changyao and Fu, Xinyu and Hu, Yuxuan and Lu, Zimu and others},
  booktitle={ACL},
  year={2026}
}

@inproceedings{viewspatial,
  title={ViewSpatial-Bench: Evaluating Multi-perspective Spatial Localization in Vision-Language Models},
  author={Li, Dingming and Li, Hongxing and Wang, Zixuan and Yan, Yuchen and Zhang, Hang and Chen, Siqi and Hou, Guiyang and Jiang, Shengpei and Zhang, Wenqi and Shen, Yongliang and others},
  booktitle={ECCV},
  year={2026}
}

@inproceedings{allangles,
  title={Seeing from another perspective: Evaluating multi-view understanding in mllms},
  author={Yeh, Chun-Hsiao and Wang, Chenyu and Tong, Shengbang and Cheng, Ta-Ying and Wang, Ruoyu and Chu, Tianzhe and Zhai, Yuexiang and Chen, Yubei and Gao, Shenghua and Ma, Yi},
  booktitle={AAAI},
  year={2026}
}

@inproceedings{ma20253dsrbench,
  title={3dsrbench: A comprehensive 3d spatial reasoning benchmark},
  author={Ma, Wufei and Chen, Haoyu and Zhang, Guofeng and Chou, Yu-Cheng and Chen, Jieneng and de Melo, Celso and Yuille, Alan},
  booktitle={ICCV},
  year={2025}
}

@inproceedings{yang2025mmsi,
  title={MMSI-Bench: A Benchmark for Multi-Image Spatial Intelligence},
  author={Yang, Sihan and Xu, Runsen and Xie, Yiman and Yang, Sizhe and Li, Mo and Lin, Jingli and Zhu, Chenming and Chen, Xiaochen and Duan, Haodong and Yue, Xiangyu and Lin, Dahua and Wang, Tai and Pang, Jiangmiao},
  booktitle={ICLR},
  year={2025}
}

@inproceedings{mindcube,
  title={Spatial mental modeling from limited views},
  author={Yin, Baiqiao and Wang, Qineng and Zhang, Pingyue and Zhang, Jianshu and Wang, Kangrui and Wang, Zihan and Zhang, Jieyu and Chandrasegaran, Keshigeyan and Liu, Han and Krishna, Ranjay and others},
  booktitle={Structural Priors for Vision Workshop at ICCV'25},
  year={2025}
}

@inproceedings{VSI,
  title={Thinking in space: How multimodal large language models see, remember, and recall spaces},
  author={Yang, Jihan and Yang, Shusheng and Gupta, Anjali W and Han, Rilyn and Fei-Fei, Li and Xie, Saining},
  booktitle={CVPR},
  year={2025}
}

@inproceedings{spar,
  title={From flatland to space: Teaching vision-language models to perceive and reason in 3d},
  author={Zhang, Jiahui and Chen, Yurui and Xu, Yueming and Huang, Ze and Mei, Jilin and Chen, Chunhui and Zhou, Yanpeng and Yuan, Yu-Jie and Cai, Xinyue and Huang, Guowei and others},
  booktitle={NeurIPS},
  year={2026}
}

@inproceedings{tong2024cambrian,
  title={Cambrian-1: A fully open, vision-centric exploration of multimodal llms},
  author={Tong, Shengbang and Brown II, Ellis L and Wu, Penghao and Woo, Sanghyun and Iyer, Adithya Jairam and Akula, Sai Charitha and Yang, Shusheng and Yang, Jihan and Middepogu, Manoj and Wang, Ziteng and others},
  booktitle={NeurIPS},
  year={2024}
}

@article{DA2K,
  title={Depth Anything V2},
  author={Yang, Lihe and Kang, Bingyi and Huang, Zilong and Zhao, Zhen and Xu, Xiaogang and Feng, Jiashi and Zhao, Hengshuang},
  journal={arXiv:2406.09414},
  year={2024}
}

@inproceedings{PGPS9K,
  title     = {A Multi-Modal Neural Geometric Solver with Textual Clauses Parsed from Diagram},
  author    = {Zhang, Ming-Liang and Yin, Fei and Liu, Cheng-Lin},
  booktitle = {IJCAI},
  year      = {2023},
}

@inproceedings{Geometry3K,
  title={Inter-gps: Interpretable geometry problem solving with formal language and symbolic reasoning},
  author={Lu, Pan and Gong, Ran and Jiang, Shibiao and Qiu, Liang and Huang, Siyuan and Liang, Xiaodan and Zhu, Song-Chun},
  booktitle={ACL},
  year={2021}
}

@article{zhang2024mathverse,
  title={Mathverse: Does your multi-modal llm truly see the diagrams in visual math problems?},
  author={Zhang, Renrui and Jiang, Dongzhi and Zhang, Yichi and Lin, Haokun and Guo, Ziyu and Qiu, Pengshuo and Zhou, Aojun and Lu, Pan and Chang, Kai-Wei and Gao, Peng and others},
  journal={arXiv preprint arXiv:2403.14624},
  year={2024}
}

@inproceedings{lu2024mathvista,
  author    = {Lu, Pan and Bansal, Hritik and Xia, Tony and Liu, Jiacheng and Li, Chunyuan and Hajishirzi, Hannaneh and Cheng, Hao and Chang, Kai-Wei and Galley, Michel and Gao, Jianfeng},
  title     = {MathVista: Evaluating Mathematical Reasoning of Foundation Models in Visual Contexts},
  booktitle={ICLR},
  year      = {2024}
}

@inproceedings{zhang2025mavis,
  title={Mavis: Mathematical visual instruction tuning with an automatic data engine},
  author={Zhang, Renrui and Wei, Xinyu and Jiang, Dongzhi and Guo, Ziyu and Zhang, Yichi and Tong, Chengzhuo and Liu, Jiaming and Zhou, Aojun and Zhang, Shanghang and Peng, Gao and others},
  booktitle={ICLR},
  year={2025}
}

@article{lin2026mmfinereason,
  title={Mmfinereason: Closing the multimodal reasoning gap via open data-centric methods},
  author={Lin, Honglin and Liu, Zheng and Zhu, Yun and Qin, Chonghan and Lin, Juekai and Shang, Xiaoran and He, Conghui and Zhang, Wentao and Wu, Lijun},
  journal={arXiv preprint arXiv:2601.21821},
  year={2026}
}

@article{wiedmann2025finevision,
  title={Finevision: Open data is all you need},
  author={Wiedmann, Luis and Zohar, Orr and Mahla, Amir and Wang, Xiaohan and Li, Rui and Frere, Thibaud and von Werra, Leandro and Gosthipaty, Aritra Roy and Marafioti, Andr{\'e}s},
  journal={arXiv preprint arXiv:2510.17269},
  year={2025}
}

@misc{numina_math_datasets,
  author = {Jia LI and Edward Beeching and Lewis Tunstall and Ben Lipkin and Roman Soletskyi and Shengyi Costa Huang and Kashif Rasul and Longhui Yu and Albert Jiang and Ziju Shen and Zihan Qin and Bin Dong and Li Zhou and Yann Fleureau and Guillaume Lample and Stanislas Polu},
  title = {NuminaMath},
  year = {2024},
  publisher = {Numina},
  journal = {Hugging Face repository},
  howpublished = {\url{[https://huggingface.co/AI-MO/NuminaMath-CoT](https://github.com/project-numina/aimo-progress-prize/blob/main/report/numina_dataset.pdf)}}
}

@article{virl39k,
      title={VL-Rethinker: Incentivizing Self-Reflection of Vision-Language Models with Reinforcement Learning},
      author = {Wang, Haozhe and Qu, Chao and Huang, Zuming and Chu, Wei and Lin,Fangzhen and Chen, Wenhu},
      journal={arXiv preprint arXiv:2504.08837},
      year={2025}
}

@inproceedings{honeybee,
  title={Bee: A high-quality corpus and full-stack suite to unlock advanced fully open mllms},
  author={Zhang, Yi and Ni, Bolin and Chen, Xin-Sheng and Zhang, Hengrui and Rao, Yongming and Peng, Houwen and Lu, Qinglin and Hu, Winston and Guo, Meng-Hao and Hu, Shi-Min},
  booktitle={ICLR},
  year={2026}
}

@inproceedings{VAE,
  title={Auto-Encoding Variational Bayes},
  author={Kingma, Diederik P and Welling, Max},
  booktitle={ICLR},
  year={2014},
}

@article{kimi3,
  title={Kimi K3: Open frontier intelligence},
  author={Team, Kimi and Bai, Tongtong and Bai, Yifan and Bao, Yiping and Cai, Jianfeng and Cai, Xinyuan and Cao, Peizhou and Cao, Yuxuan and Chai, Ziwei and Charles, Y and others},
  journal={arXiv preprint arXiv:2607.24653},
  year={2026}
}

@article{han2026towards,
  title={Towards Physics of Multimodal Pretraining: Knowledge Flow, Modality Synergy, Early Unification, and Recipes},
  author={Han, Junlin and Tong, Shengbang and Fan, David and Chen, Minghao and Torr, Philip and Kokkinos, Filippos and Lewis, Mike},
  journal={arXiv preprint arXiv:2608.05000},
  year={2026}
}

@article{tong2026beyond,
  title={Beyond language modeling: An exploration of multimodal pretraining},
  author={Tong, Shengbang and Fan, David and Nguyen, John and Brown, Ellis and Zhou, Gaoyue and Qian, Shengyi and Zheng, Boyang and Vallaeys, Th{\'e}ophane and Han, Junlin and Fergus, Rob and others},
  journal={arXiv preprint arXiv:2603.03276},
  year={2026}
}

@article{thinkingmachines2026interactionmodels,
  author  = {Thinking Machines Lab},
  title   = {Interaction Models: A Scalable Approach to Human-AI Collaboration},
  journal = {Thinking Machines Lab: Connectionism},
  year    = {2026},
  month   = {May},
  note    = {https://thinkingmachines.ai/blog/interaction-models/},
  doi     = {10.64434/tml.20260511},
}

@misc{fuyu-8b,
  author = {Bavishi, Rohan and Elsen, Erich and Hawthorne, Curtis and Nye, Maxwell and Odena, Augustus and Somani, Arushi and  Ta\c{s}\i{}rlar, Sa\u{g}nak},
  title = {Introducing our Multimodal Models},
  url = {https://www.adept.ai/blog/fuyu-8b},
  year = {2023}
}

@article{uno,
  title={Steering Visual Generation in Unified Multimodal Models with Understanding Supervision},
  author={Liu, Zeyu and Ni, Zanlin and Yue, Yang and Da, Cheng and Yang, Huan and Zhang, Di and Gai, Kun and Huang, Gao},
  journal={arXiv preprint arXiv:2605.05781},
  year={2026}
}

@article{kang2026transferability,
  title={Transferability Between Understanding and Generation in Unified Multimodal Models},
  author={Kang, Jiwon and Yoon, Heeji and Jung, Jaewoo and Min, Jaewon and Jeon, Minkyeong and Hwang, Biyeon and Jung, Sangwon and Kim, Seungryong},
  journal={arXiv preprint arXiv:2607.04423},
  year={2026}
}

@inproceedings{mrope,
  title={Revisiting multimodal positional encoding in vision--language models},
  author={Huang, Jie and Liu, Xuejing and Song, Sibo and Hou, Ruibing and Chang, Hong and Lin, Junyang and Bai, Shuai},
  booktitle={ICLR},
  year={2026}
}

@article{gabeur2026imagegenerators,
  title={Image generators are generalist vision learners},
  author={Gabeur, Valentin and Long, Shangbang and Peng, Songyou and Voigtlaender, Paul and Sun, Shuyang and Bao, Yanan and Truong, Karen and Wang, Zhicheng and Zhou, Wenlei and Barron, Jonathan T and others},
  journal={arXiv preprint arXiv:2604.20329},
  year={2026}
}

@article{wang2026videogeneration,
  title={Video Generation Models are General-Purpose Vision Learners},
  author={Wang, Letian and Zhang, Chuhan and Kabra, Rishabh and Uijlings, Jasper and Waslander, Steven and Zisserman, Andrew and Carreira, Joao and He, Kaiming and Andriluka, Misha and Bazavan, Eduard Gabriel and others},
  journal={arXiv preprint arXiv:2607.09024},
  year={2026}
}

@inproceedings{LPIPS,
  author={Zhang, Richard and Isola, Phillip and Efros, Alexei A. and Shechtman, Eli and Wang, Oliver},
  booktitle={CVPR}, 
  title={The Unreasonable Effectiveness of Deep Features as a Perceptual Metric}, 
  year={2018}}

@article{SSIM,
  author={Zhou Wang and Bovik, A.C. and Sheikh, H.R. and Simoncelli, E.P.},
  journal={IEEE Transactions on Image Processing}, 
  title={Image quality assessment: from error visibility to structural similarity}, 
  year={2004},}

@article{lin2026perceptionbench,
  title={PerceptionBench: Evaluating Atomic Visual Perception in Multimodal Large Language Models},
  author={Lin, Zichao and Xie, Yifeng and Qu, Bowen and Wang, Haiming and Li, Jia and Wu, Haoning and Dong, Yuhao and Yang, Zuhao and Zhu, Jinguo and Lu, Haoyu and others},
  journal={arXiv preprint arXiv:2607.24957},
  year={2026}
}

@inproceedings{fu2024blink,
  title={Blink: Multimodal large language models can see but not perceive},
  author={Fu, Xingyu and Hu, Yushi and Li, Bangzheng and Feng, Yu and Wang, Haoyu and Lin, Xudong and Roth, Dan and Smith, Noah A and Ma, Wei-Chiu and Krishna, Ranjay},
  booktitle={ECCV},
  year={2024},
}

@article{lars,
  title={Large batch training of convolutional networks},
  author={You, Yang and Gitman, Igor and Ginsburg, Boris},
  journal={arXiv preprint arXiv:1708.03888},
  year={2017}
}

@misc{openai2026gpt56sol,
  title        = {{GPT-5.6}: Frontier Intelligence that Scales with Your Ambition},
  author       = {{OpenAI}},
  year         = {2026},
  url          = {https://openai.com/index/gpt-5-6/},
  note         = {Large language model}
}
\bibliographystyle{iclr2027_conference}

\clearpage
\appendix

\section{Representation-level Details}
\label{app:representation_details}

\subsection{Data}
\label{app:representation_details_data}
\begin{figure}[h]
  \centering
  \includegraphics[width=0.8\linewidth]{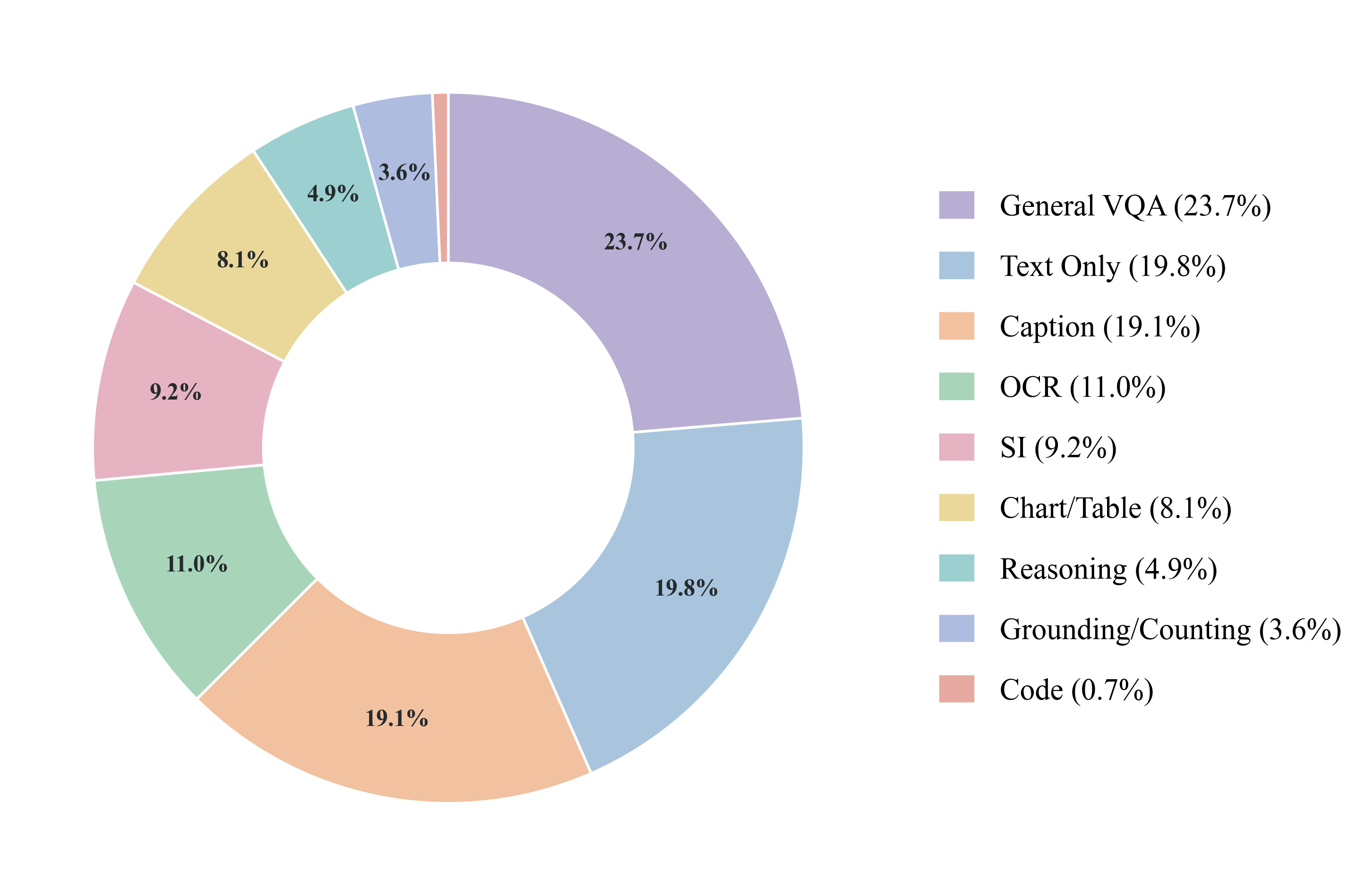}
  \caption{Category distribution of the understanding training
  data.}
  \label{fig:representation-data-distribution}
\end{figure}

We draw the training data from SenseNova-U1 using a tailored sampling mixture. The
category distribution of the visual-understanding data is summarized in
\paperfigref{fig:representation-data-distribution}. For visual generation, we exclude infographic data and downweight
samples that emphasize text rendering. We only include generation samples with
English prompts.
For joint training, visual-understanding, text-to-image, and image-editing data
are sampled with a relative ratio of $3:6:1$. The task-decoupled \mot{}
additionally includes two parts of reconstruction data. For fair comparison, we keep the batch sizes for understanding and
generation data identical between the joint models and their corresponding
\und{}-only and \gen{}-only counterparts. Training for 210K steps exposes each
model to approximately 78M understanding samples, 159M text-to-image samples,
and 26.5M image-editing samples. For the 3D spatial-intelligence case study, we
retrain the base model after removing all SI-related examples, preventing data overlap and isolating transfer from the
task-specific data introduced in that study.

\subsection{Model}
\label{app:representation_details_model}

All models are initialized from Qwen3-1.7B with a 16-layer pre-buffer. We use an MLP-based timestep embedder and noise scale embedder to extract embeddings and directly add them to the visual inputs, similar to SenseNova-U1. Different from SenseNova-U1, we do not have additional norm layers for visual hidden states as we choose a different multimodal positional embedding scheme.

\subsection{Training}
\label{app:representation_details_training}
\begin{table}[h]
  \caption{Representation-level training configuration.}
  \label{tab:training-details}
  \centering
  \small
  \begin{tabular}{ll}
    \toprule
    Item & Value \\
    \midrule
    Peak learning rate & $1\times10^{-4}$ \\
    Learning-rate scheduler & Constant \\
    Optimizer & AdamW ($\beta_1$ = 0.9, $\beta_2$ = 0.95, $\epsilon$ = $10^{-8}$) \\
    Weight decay & 0.0 \\
    Gradient-norm clipping & 1.0 \\
    EMA ratio & 0.9999 \\
    Training steps & 210K \\
    Warmup steps & 2000 \\
    Loss weight (CE : MSE) & 0.1:1 \\
    Understanding resolution & $[256^2, 2048^2]$ \\
    Generation resolution & $[512^2, 1024^2]$ \\
    Sequence length & 16K \\
    Timestep shift & $\mu = -0.8, \sigma = 0.8$ in the logit-normal t-sampler \\
    Noise scale & $\sqrt{N/64}$, where $N$ is \# of tokens \\
    \bottomrule
  \end{tabular}
\end{table}
Training details in our representation-level studies are provided in \papertabref{tab:training-details}.

\subsection{Evaluation}
\label{app:representation_details_evaluation}

\begin{table}[h]
  \caption{Full results for visual understanding benchmarks. Higher is better for all metrics.}
  \label{tab:representation-full-results}
  \centering
  \scriptsize
  \setlength{\tabcolsep}{4pt}
  \resizebox{\textwidth}{!}{%
  \begin{tabular}{llccccc}
    \toprule
    Group & Benchmark & Dense-U & Dense-U+G
      & Mod.-dec. MoT U & Mod.-dec. MoT U+G
      & Task-dec. MoT U+G \\
    \midrule
    \multirow{5}{*}{General}
      & MME & 1930.46 & 1967.90 & 1773.31 & 1666.54 & 1992.06 \\
      & MMBench & 71.39 & 69.58 & 64.43 & 66.66 & 69.84 \\
      & MMStar & 52.20 & 55.82 & 50.72 & 47.69 & 56.51 \\
      & SEED Bench-I & 76.56 & 76.91 & 74.33 & 71.91 & 76.42 \\
      & MMMU & 42.44 & 42.77 & 42.89 & 38.33 & 42.77 \\
    \midrule
    \multirow{5}{*}{OCR}
      & DocVQA & 94.37 & 93.13 & 90.05 & 85.29 & 94.63 \\
      & ChartQA & 45.40 & 46.84 & 44.68 & 40.20 & 45.24 \\
      & InfoVQA & 68.75 & 67.52 & 52.65 & 44.59 & 68.30 \\
      & OCRBench & 67.50 & 68.90 & 63.90 & 59.90 & 67.80 \\
      & AI2D & 83.80 & 84.20 & 82.90 & 80.60 & 84.50 \\
    \midrule
    \multirow{9}{*}{V-Centric \& SI}
      & PerceptionBench$_\textrm{mcq}$ & 31.30 & 32.59 & 29.66 & 29.66 & 34.02 \\
      & P2GB & 72.63 & 73.94 & 66.40 & 63.40 & 72.07 \\
      & MME-RealWorld & 48.93 & 49.81 & 46.63 & 41.01 & 50.39 \\
      & BLINK & 63.92 & 64.09 & 60.07 & 59.02 & 63.48 \\
      & DA-2K & 74.85 & 74.85 & 77.22 & 72.24 & 77.46 \\
      & CV-Bench & 81.88 & 81.27 & 79.15 & 79.90 & 82.44 \\
      & MindCube & 58.57 & 61.81 & 62.67 & 58.19 & 64.85 \\
      & 3DSR & 59.01 & 61.78 & 59.78 & 59.14 & 61.02 \\
      & ViewSpatial & 55.41 & 55.86 & 54.32 & 54.90 & 55.75 \\
    \bottomrule
  \end{tabular}
  }
\end{table}

\begin{table*}[t]
  \centering
  \begin{minipage}[t]{0.49\textwidth}
    \centering
    \captionof{table}{GenEval2 results by skill. We report atom-level scores;
    higher is better.}
    \label{tab:geneval2-breakdown}
    \setlength{\tabcolsep}{3.2pt}
    \small
    \resizebox{\linewidth}{!}{%
    \begin{tabular}{lcccccc}
      \toprule
      Model & Object & Attribute & Count & Position & Verb & \textbf{Overall} \\
      \midrule
      Dense-G
        & 67.91 & 56.08 & 34.34 & 33.35 & 9.02 & 51.16 \\
      Dense-U+G
        & 67.08 & 53.97 & 34.61 & 33.67 & 7.45 & 51.03 \\
      Mod.-dec. \mot{}-G
        & 77.65 & 58.69 & 39.76 & 42.81 & 13.75 & 57.55 \\
      Mod.-dec. \mot{}-U+G
        & 84.60 & 63.26 & 43.95 & 53.05 & 9.28 & 63.47 \\
      Task-dec. \mot{}-U+G
        & 82.97 & 64.45 & 45.6 & 56.11 & 15.77 & 63.96 \\
      \bottomrule
    \end{tabular}%
    }
  \end{minipage}
  \hfill
  \begin{minipage}[t]{0.49\textwidth}
    \centering
    \captionof{table}{DPG-Bench results by level-1 category. Higher is
    better.}
    \label{tab:dpg-breakdown}
    \setlength{\tabcolsep}{3.2pt}
    \small
    \resizebox{\linewidth}{!}{%
    \begin{tabular}{lcccccc}
      \toprule
      Model & Global & Entity & Attribute & Relation & Other & \textbf{Overall} \\
      \midrule
      Dense-G
        & 91.95 & 85.55 & 88.83 & 87.52 & 79.49 & 79.11 \\
      Dense-U+G
        & 83.58 & 85.77 & 87.25 & 89.45 & 84.63 & 78.12 \\
      Mod.-dec. \mot{}-G
        & 88.02 & 89.82 & 88.58 & 90.42 & 89.20 & 82.06 \\
      Mod.-dec. \mot{}-U+G
        & 92.14 & 88.12 & 90.39 & 92.03 & 90.77 & 83.09 \\
      Task-dec. \mot{}-U+G
        & 87.30 & 90.50 & 89.09 & 90.65 & 91.17 & 82.31 \\
      \bottomrule
    \end{tabular}%
    }
  \end{minipage}
\end{table*}

The full per-benchmark results for visual understanding benchmarks are provided in \papertabref{tab:representation-full-results}. We use the image subset for SEED-Bench, the lite subset for MME-RealWorld, and the mini subset for MindCube. We only use the MCQ format questions in PerceptionBench for more stable evaluation. All benchmarks are evaluated with temperature set to 0. Detailed results by skill and level-1 category for GenEval2 and DPG-Bench are reported in \papertabref{tab:geneval2-breakdown} and \papertabref{tab:dpg-breakdown}, respectively.

For generation-related evaluation, we generate images at a resolution of
$1024\times1024$ using a classifier-free guidance scale of 4.0, 50 sampling
steps, and a timestep shift of 3.0.

\subsection{Probing Experiments}
\label{app:representation_details_probing}

For ImageNet classification, we follow the standard linear-probing protocol.
We freeze the multimodal backbone, globally pool the visual features, and
train a linear classification head using the LARS \citep{lars} optimizer for 100 epochs.

For semantic segmentation and monocular depth estimation, we conduct dense
probing with the backbone similarly frozen. Because the model produces visual
features at a relatively large spatial downsampling factor of 32, the probing
head contains three consecutive convolution-and-upsampling blocks to recover
spatial resolution. We train the segmentation and depth heads for 30 epochs,
optimizing only the parameters of the corresponding probing head.

\section{Task-level Details}
\label{app:tasks_details}

\subsection{SVG Diagnostics}
\label{app:tasks_details_svg}

\begin{figure*}[p]
  \centering
  \includegraphics[width=\textwidth,height=0.82\textheight,keepaspectratio]{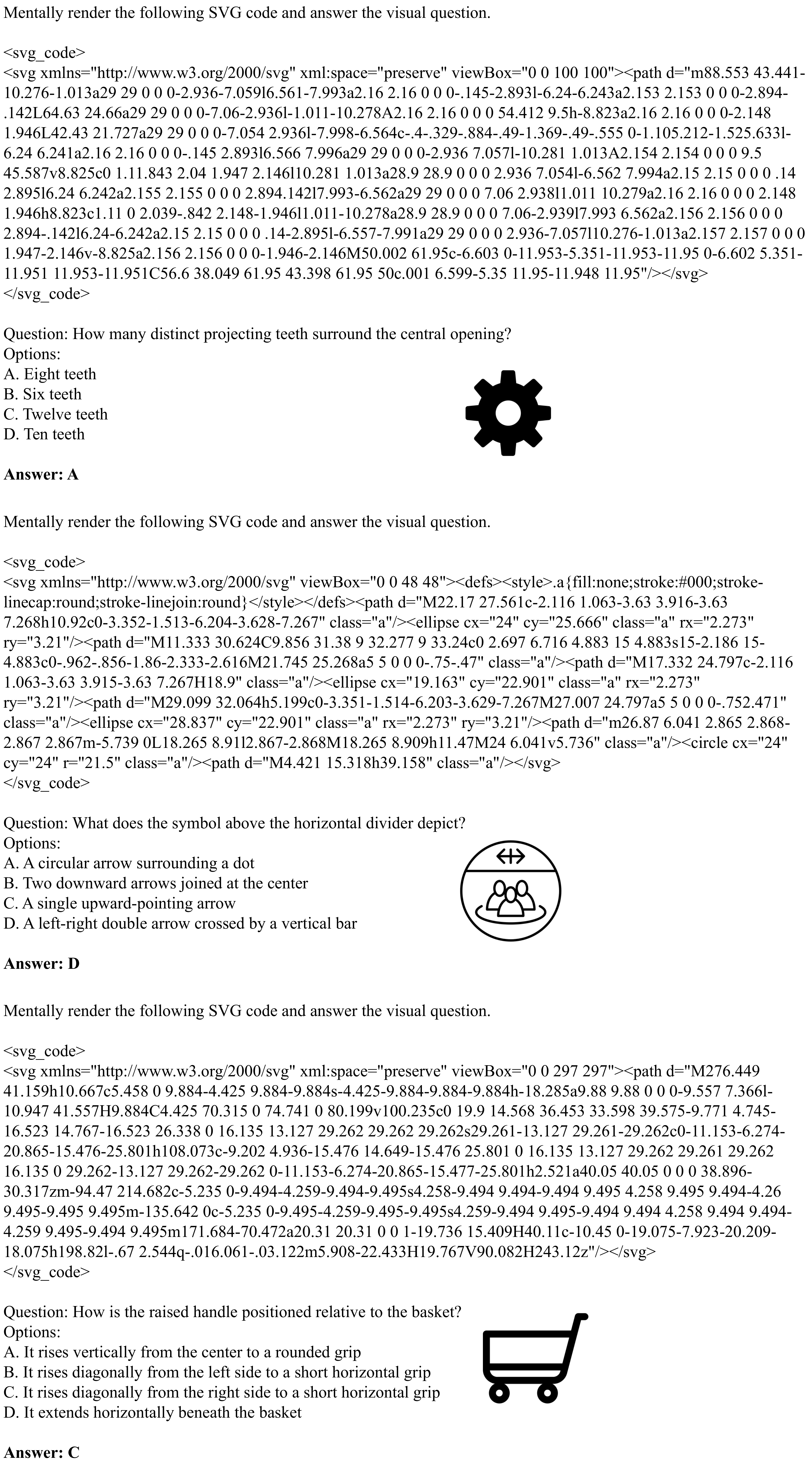}
  \caption{\textbf{Examples from the SVG Blind VQA diagnostic.} The model is
  given only the SVG code, question, and answer options, and must infer the
  visual answer without observing the rendered icon. The renderings shown here
  are provided solely for readers.}
  \label{fig:svg-blind-vqa-examples}
\end{figure*}

We construct an SVG Blind VQA diagnostic to evaluate whether the model can
infer visual appearance directly from SVG code. Starting from the UniSVG test
set, we use GPT-5.6-luna \citep{openai2026gpt56sol} to annotate 3,000
multiple-choice VQA examples. To retain nontrivial questions, we evaluate them
with our base model, which has not been trained on SVG-specific data, and keep
only the examples it answers incorrectly. This filtering yields 1,370
examples. Examples are shown in
\paperfigref{fig:svg-blind-vqa-examples}.

\subsection{Generation Tasks for 3D SI}
\label{app:tasks_details_si}
\begin{figure*}[p]
  \centering
  \textbf{(a) Ego-motion Transition}\\[3pt]
  \includegraphics[width=0.98\textwidth]{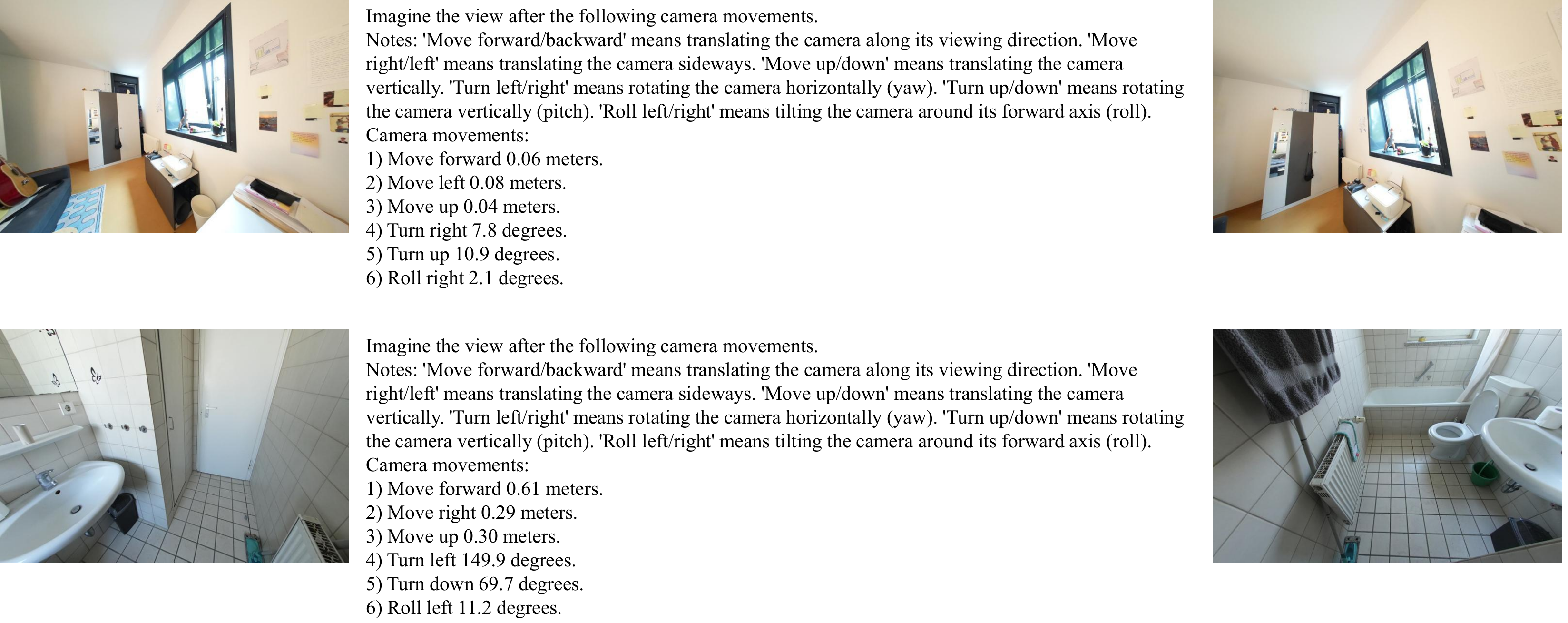}

  \vspace{8pt}
  \textbf{(b) Multi-view Reconstruction}\\[3pt]
  \includegraphics[width=0.98\textwidth]{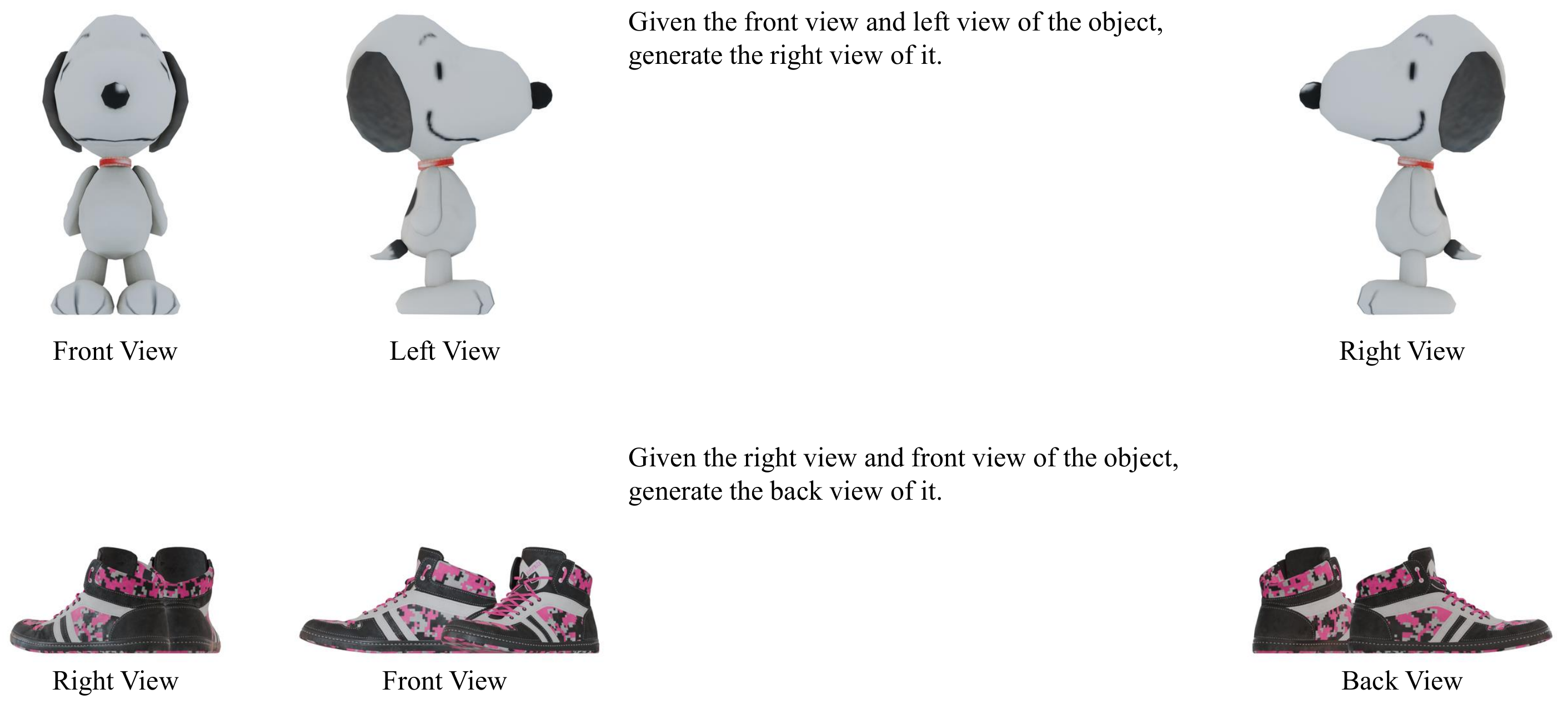}
  \caption{\textbf{Examples of 3D spatial-intelligence generation tasks
  (I).} Ego-motion transition requires applying explicit camera motion to a
  source view, while multi-view reconstruction infers an unobserved view from
  the provided object views.}
  \label{fig:3d-generation-examples-a}
\end{figure*}

\begin{figure*}[p]
  \centering
  \textbf{(c) View Sequence Completion}\\[3pt]
  \includegraphics[width=0.98\textwidth]{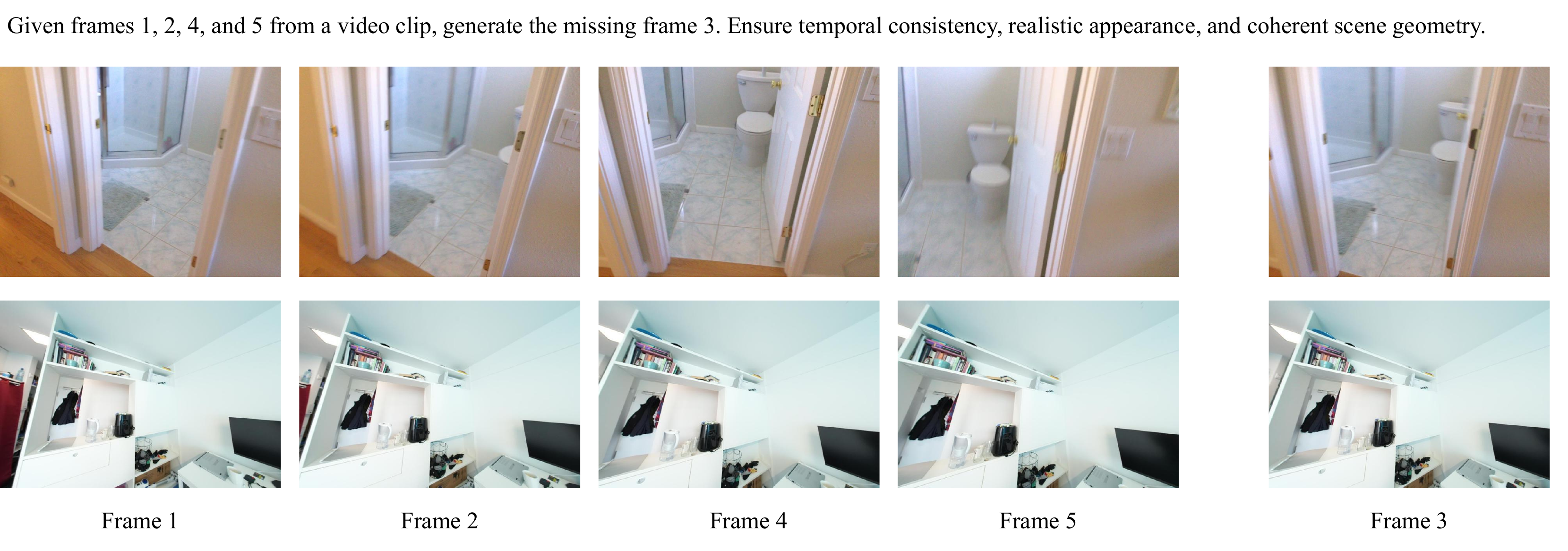}

  \vspace{8pt}
  \textbf{(d) Layout-to-image Generation}\\[3pt]
  \includegraphics[width=0.98\textwidth]{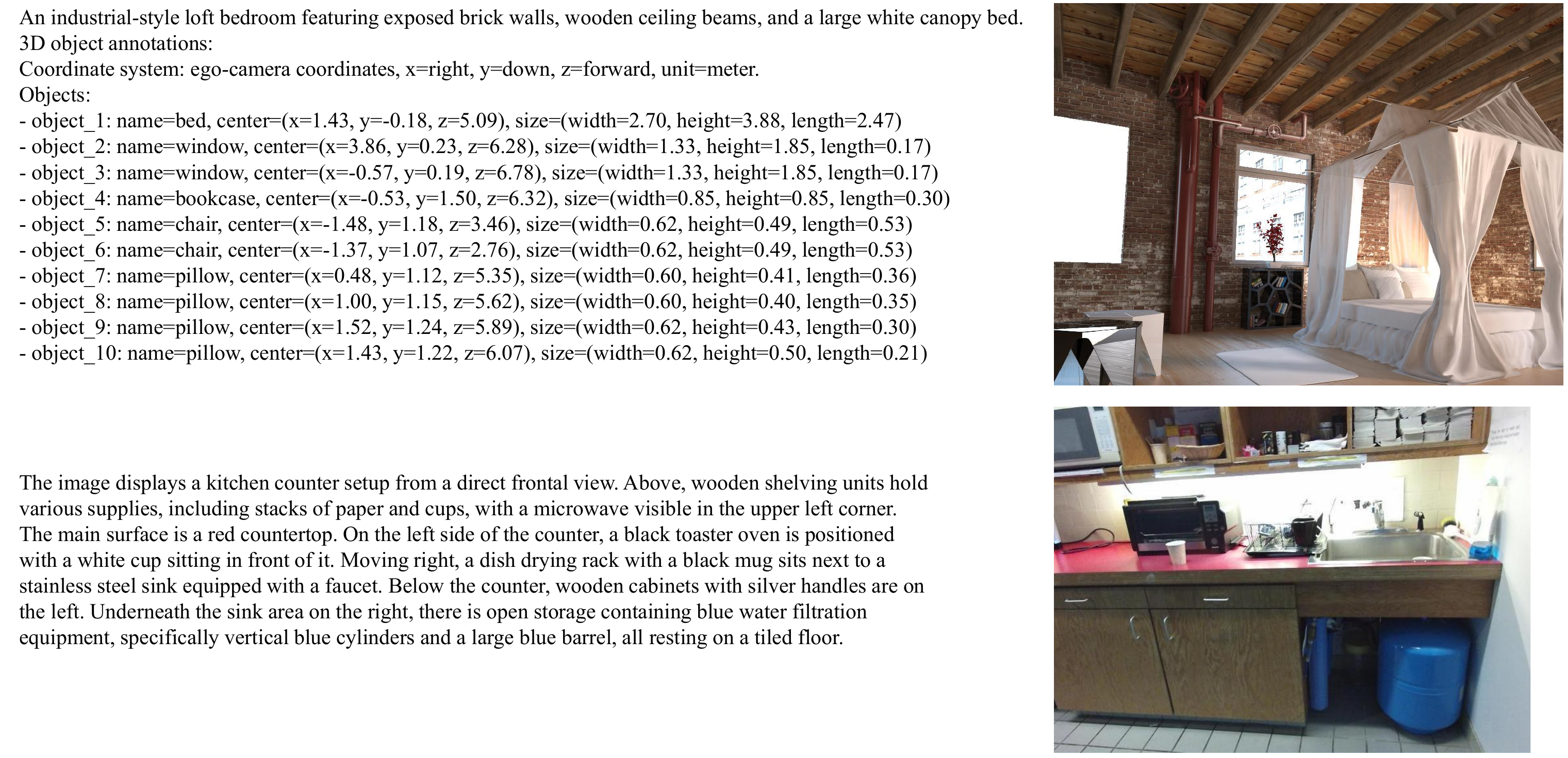}
  \caption{\textbf{Examples of 3D spatial-intelligence generation tasks
  (II).} View sequence completion predicts a temporally and geometrically
  consistent missing frame, while layout-to-image generation synthesizes an
  image from structured or textual descriptions of 3D scene layout.}
  \label{fig:3d-generation-examples-b}
\end{figure*}

We construct four generation tasks that require different aspects of 3D
spatial intelligence. Ego-motion transition predicts a target view from a
source image and a sequence of camera transformations. Multi-view
reconstruction generates an unobserved object view from two provided views.
View sequence completion recovers the missing middle frame of a short camera
trajectory. Layout-to-image generation renders a scene from either structured
3D object annotations or a natural-language description of its spatial
layout. Examples of the four task formats are shown in
\paperfigref{fig:3d-generation-examples-a} and
\paperfigref{fig:3d-generation-examples-b}.

\end{document}